\documentclass[journal]{IEEEtran} 

\usepackage{amsmath,amssymb,amsfonts,bm,mathdots,amsthm}

\usepackage{graphicx}
\usepackage{float}
\usepackage{wrapfig}
\usepackage{rotating}
\usepackage{placeins}

\usepackage{array}
\usepackage{multirow}
\usepackage{booktabs}
\usepackage{longtable}
\usepackage{adjustbox}
\usepackage{tabularx}
\usepackage{tabularray}
\usepackage{makecell}
\usepackage{url}
\usepackage{hyperref}

\usepackage{algorithm}
\usepackage{algorithmic}

\usepackage{caption}
\usepackage{subcaption}
\usepackage{subcaption}

\usepackage{textcomp}
\usepackage{ragged2e}
\usepackage{multicol}
\usepackage{cuted}
\usepackage{etoolbox}
\AfterEndEnvironment{strip}{\leavevmode}
\usepackage{xcolor}
\usepackage{colortbl}
\definecolor{lightblue}{RGB}{173,216,230}
\definecolor{skyblue}{RGB}{0,119,182}

\usepackage{cite}
\usepackage[utf8]{inputenc}
\usepackage{csquotes}

\usepackage[utf8]{inputenc}

\usepackage{enumitem}
\usepackage{pifont}
\usepackage{nidanfloat}

\newcommand{\cmark}{\ding{51}} 
\newcommand{\xmark}{\ding{55}} 

\def\BibTeX{{\rm B\kern-.05em{\sc i\kern-.025em b}\kern-.08em
  T\kern-.1667em\lower.7ex\hbox{E}\kern-.125emX}}

\begin{document}
\title{Robust and Real-Time Bangladeshi Currency Recognition: A Dual-Stream MobileNet and EfficientNet Approach}

\author{
    \IEEEauthorblockN{
        Subreena\textsuperscript{1},
        Mohammad Amzad Hossain\textsuperscript{1},
        Mirza Raquib\textsuperscript{1,2},  
        Saydul Akbar Murad\textsuperscript{3}, 
        Farida Siddiqi Prity\textsuperscript{4}, 
        Muhammad Hanif\textsuperscript{1},
        Nick Rahimi\textsuperscript{3,*}
    }

    \IEEEauthorblockA{
        \textsuperscript{1}Department of Information and Communication Engineering, Noakhali Science and Technology University, Noakhali, Bangladesh\\
        \textsuperscript{2}Department of Computer and Communication Engineering, International Islamic University Chittagong, Chattogram, Bangladesh\\
        \textsuperscript{3}School of Computing Sciences and Computer Engineering, University of Southern Mississippi, Hattiesburg, MS, USA\\
        \textsuperscript{4}Department of CSE, Netrokona University, Netrokona, Bangladesh\\
        Emails: \{subreena1115, amzad, drhanifmurad\}@nstu.edu.bd, mirzaraquib@iiuc.ac.bd, \\ 
        \{saydulakbar.murad, nick.rahimi\}@usm.edu, prity@neu.ac.bd
    }
}


\maketitle

\begin{abstract}

Accurate currency recognition is essential for assistive technologies, particularly for visually impaired individuals who rely on others to identify banknotes. This dependency puts them at risk of fraud and exploitation. To address these challenges, we first build a new Bangladeshi banknote dataset that includes both controlled and real-world scenarios, ensuring a more comprehensive and diverse representation. Next, to enhance the dataset's robustness, we incorporate four additional datasets, including public benchmarks, to cover various complexities and improve the model’s generalization. To overcome the limitations of current recognition models, we propose a novel hybrid CNN architecture that combines MobileNetV3-Large and EfficientNetB0 for efficient feature extraction. This is followed by an effective multilayer perceptron (MLP) classifier to improve performance while keeping computational costs low, making the system suitable for resource-constrained devices. The experimental results show that the proposed model achieves 97.95\% accuracy on controlled datasets, 92.84\% on complex backgrounds, and 94.98\% accuracy when combining all datasets. The model’s performance is thoroughly evaluated using five-fold cross-validation and seven metrics: accuracy, precision, recall, F1-score, Cohen’s Kappa, MCC, and AUC. Additionally, explainable AI methods like LIME and SHAP are incorporated to enhance transparency and interpretability.  The code for this research can be found in this link: \href{https://github.com/subreena/bangladeshi\_banknote\_recognition/blob/main/banknote\_recongnition.ipynb}{https://github.com/subreena/bangladeshi}


\end{abstract} 

\section{Introduction}
\label{sec:introduction}

\IEEEPARstart{T}{he} proliferation of technology has enhanced life quality for many, notably in health, education, and communication domains, and assistive technologies have been instrumental in empowering individuals with disabilities, such as the visually impaired \cite{1kola2023life}, \cite{ VIPassist1}, \cite{VIPassist2}. Visual impairment significantly limits independence during routine financial transactions, particularly in identifying currency, impacting around 2.2 billion individuals globally, with a concentration in lower-income nations \cite{3world_2023}. In Bangladesh, approximately 650,000 adults are visually impaired, with a reported prevalence of blindness of approximately 1 in individuals aged 40 and older \cite{bd_blindness2003prevalence},\cite{bd_blindness2022prevalence}. While tactile identification is possible with Bangladeshi currency notes due to surface features like raised dots \cite{intaglio_tactile}, \cite{bd_tactile_wiki}, \cite{ bd_tactile_web}. These features deteriorate over time and are not consistently reliable for denomination recognition, given the comparable dimensions, textures, and visual designs of different notes. This situation can threaten financial independence, particularly for younger people and the elderly. Therefore, there's an urgent need for a cost-effective, easily accessible, and reliable identification system for the main currency used in cash-based, rapidly growing South Asian economies.

Traditional methods for recognizing currency have relied on tactile methods developed for visually impaired people and have consisted of identifying the denomination by size and texture \cite{tactile_semary2015currency}. Other approaches include seeking human assistance or specialized electronic devices, which are limited by poor accuracy, accessibility, and trust \cite{tactile_yousry2018currency}. While deep learning-based systems for currency recognition showed promising outcomes in controlled environments, their deployment in resource-constrained environments has limitations, including computational overhead limiting real-time low-resource device implementation, poor generalization to diverse real-world conditions, including complex backgrounds and varying lighting, insufficient robustness to worn or partially occluded banknotes, and a lack of interpretability, which can undermine user trust and obstruct system debugging.
Recent advances in deep learning architectures have opened new possibilities for mobile computer vision applications, achieving optimal trade-offs between model size and accuracy through techniques such as depthwise separable convolutions and compound scaling \cite{MobileNetV3}, \cite{tan2019efficientnet}. Researchers are using various algorithms like FAST, SIFT, ORB, and SURF \cite{9ganjavecurrency}, \cite{10roy2017development}, \cite{Sarker2019}  and deep learning models like CNN, YOLO, Inception, MobileNet, EfficientNet, NasNet, ResNet \cite{ 12linkon2020deep}, \cite{ 13joshi2020yolov3}, \cite{8singh2022ipcrf}, \cite{MobileApp1Pujari},  to find a better approach to help the visually impaired individuals in independent financial transactions. Furthermore, hybrid deep learning architectures that combine the strengths of different models have shown robust performance in various computer vision tasks, including medical imaging, remote sensing, IoT security, and visual recognition \cite{hybrid1_nazir2024deep}, \cite{hybrid2_yang2025enhanced}, \cite{hybrid3_terzioglu2024optimizing}, \cite{hybrid4_ye2025hybrid}. Most existing Bangladeshi currency recognition systems predominantly focus on single-architecture approaches assessed on limited datasets and fail to address the complexity encountered in practical deployment scenarios \cite{12linkon2020deep}, \cite{7tasnim2021bangladeshi}. Moreover, the black-box nature of deep learning models poses critical challenges for assistive technologies, where transparency and explainable decisions are essential to ensuring user trust. Despite its significance, the adoption of explainable AI techniques such as LIME (Local Interpretable Model-agnostic Explanations) and SHAP (Shapley Additive exPlanations) remains minimal \cite{xai_ribeiro2016should}, \cite{xai_lundberg2017unified}.

This paper addresses these challenges through a comprehensive investigation of hybrid deep learning architectures for real-time Bangladeshi banknote recognition. As datasets for Bangladeshi currency have grown in size and diversity, new challenges have emerged, including handling worn, occluded, folded, or counterfeit notes while ensuring accessibility and reliability for all users. The major contributions of this work are summarized as follows:

\begin{itemize}
    \item We have proposed a novel hybrid model that combines MobileNetV3-Large and EfficientNetB0 as complementary feature extractors, followed by a multi-layer perceptron (MLP) pipeline for classification.
    
    \item We have curated five datasets of increasing complexity that range from simple controlled backgrounds to highly diverse, aggregated, and complex scenarios. This facilitates a methodical assessment of model efficacy across many real-world scenarios.

    \item We have conducted a comparison of four optimization algorithms (Adam, AdamW, RMSProp, and SGD) across four learning rates (0.0001, 0.001, 0.01, and 0.1) on all five datasets to aid in guiding training strategy selection. 
    
    \item We employed a five-fold cross-validation methodology to assess the proposed model across cross multiple data partitions, measuring performance using seven metrics: Accuracy, Precision, Recall, F1-Score, Cohen’s Kappa, Matthews Correlation Coefficient (MCC), and Area Under the Curve (AUC). 
    
    \item We have incorporated explainable AI (XAI) methods, LIME and SHAP, that offer insights into the model's decision-making process that led to the predictions.
    
    \item  We have developed a practical, real-time, web-based currency recognition system with a voice-activation feature, which can generate both auditory and textual outputs, demonstrating the system's adaptability for deployment on resource-constrained devices.
    
\end{itemize}

The rest of this paper is organized as follows. Section \ref{section:related_works} reviews the related works in currency recognition. The methodology of the proposed hybrid model is explained in Section~\ref{section:methodology}, including dataset building, model architecture design, and experimental setup, along with the evaluation framework. Section~\ref{section:results} presents the comprehensive results and analysis along with a comparison to the state-of-the-art methods. Finally, the conclusion is drawn in Section~\ref{section:conclusion}.

\section{Related Works}
\label{section:related_works}

Currency recognition for visually impaired individuals has evolved as an increasingly important research domain at the intersection of assistive technology and computer vision. This domain has gradually shifted from depending on traditional image processing techniques to deep-learning models, with increasing emphasis on real-world deployment and model interpretability. Traditional image processing approaches like template matching and edge detection were used predominantly in early currency recognition systems. Prior studies on Bangladeshi currency recognition employed supervised learning approaches to recognize worn, torn, and degraded banknotes \cite{Akter2018}, highlighting the role of feature extraction for reliable recognition. Subsequent work advanced further to real-time mobile applications using image processing algorithms \cite{Sarker2019}. These studies demonstrated feasibility but lacked robustness to illumination changes, changes in note orientation, and complex background conditions, limiting their reliability in real-world usage. 
Deep learning architectures have significantly progressed currency recognition by enabling automatic extraction of discriminative features. For Bangladeshi banknotes, the work in  \cite{12linkon2020deep} explored with ResNet152v2, MobileNet, and NASNetMobile across combined datasets, achieving high accuracy on individual sets \cite{Banglacurrency2019}, \cite{BanglaMoney2018}. The study reported 98.88\% accuracy with MobileNet and 100\% with NASNetMobile, but reduced accuracy of 97.77\% on merged data, highlighting generalization challenges. Object detection approaches have also gained prominence with \cite{13joshi2020yolov3} that proposes a YOLOv3-based banknote detection and recognition model for Indian currency, achieving 95.71\% detection, 100\% recognition accuracy, while  \cite{PasumarthyYOLOV52022} used a YOLOv5-based deep neural network designed to detect Indian currency, which was trained on an annotated and augmented dataset consisting of 10,000 images. A subsequent study \cite{8singh2022ipcrf} introduced IPCRNet, a lightweight domain-specific CNN network built with dense connections and depthwise separable convolutions for deployment in resource-constrained environments, achieving 98.38\% accuracy on a large-scale dataset of over 50,000 images of Indian currency. For Bangladeshi currency, \cite{7tasnim2021bangladeshi} presented a CNN model trained on 70,000 images of Bangladeshi currency across eight denominations, reporting 92\% accuracy with textual and auditory output. However, earlier datasets of Bangladeshi currency were taken against controlled, simple background and did not contain images of 200 taka notes, as they were released much later in 2020 \cite{200tk_release}. The datasets lacked real-world environmental settings. More recently, the BanglaNotes dataset \cite{BanglaTaka2025} introduced 5,073 images across nine denominations, demonstrating emphasis towards dataset quality. Another study \cite{Siddiki2023} explored currency recognition and fraud detection using deep learning and feature extraction, though its evaluation was limited to 50 test samples. Transfer learning techniques were applied using VGG16, VGG19, InceptionV3, and ResNetV2 in \cite{Chowdhury2024} for cross-dataset banknote classification and proposed a minimalist CNN model to reduce computational cost. Similarly, a CNN-based currency recognition model was employed in the study \cite{Das2023CNN} to support visually impaired people, but lacked sufficient assessment under real-world conditions.

A multinational study \cite{park2023mbdm} introduced MBDM with a mosaic augmentation technique for Korean, US, Euro, and Jordanian currencies, achieving 93.96\% accuracy but showing a performance degradation on images with complex backgrounds and folded notes.  Recent work~\cite{Nasir2024} includes a VGG16-based recognition system for Pakistani currency, and ~\cite{14IslamViT} leveraged a Vision Transformer for Bangladeshi currency, including the 200-taka denomination, reaching 99.93\% accuracy on the Augmented Bangla Money dataset ~\cite{AugmentedBanglaMoney}. International researches include ~\cite{Ali2024Egyptian}, that used YOLOv10 for Egyptian currency, achieved 96.78\% precision, 99.34\% mAP@0.5 on 2,000 images, the study in \cite{Evwiekpaefe2024} implemented MobileNetV2 for a dataset of 3,615 images of Nigerian notes, \cite{Neto2023Brazilian} proposed a CNN-based system for Brazilian banknote recognition aimed for visually impaired people, \cite{Awad2022Iraqi} presented a deep-learning Iraqi banknote classification system to support visually impaired people in financial transactions and the authors in \cite{Bolanos2024Colombian} developed a mobile application with audio feedback to help visually impaired people recognize Colombian currency.

\begin{table*}[htbp]
\caption{Comparison of related studies with the proposed hybrid model.}
\label{tab:related}
\adjustbox{width=\textwidth,center}{
\begin{tabular}{|p{2cm}|c|c|c|c|c|c|p{3cm}|c|c|c|c|c|c|c|c|p{5cm}|}
\hline
\textbf{Study} & \multicolumn{5}{c|}{\textbf{Dataset}} & \textbf{Classes} & \textbf{Model Used} & \multicolumn{8}{c|}{\textbf{Metrics}} & \textbf{Limitations} \\
\cline{2-6} \cline{9-16}
& \rotatebox{90}{\textbf{Simple}} & \rotatebox{90}{\textbf{Complex}} & \rotatebox{90}{\textbf{Mix}} & \rotatebox{90}{\textbf{Primary}} & \rotatebox{90}{\textbf{Augmented }} 
& & 
& \rotatebox{90}{\textbf{Acc.}} & \rotatebox{90}{\textbf{Pre.}} & \rotatebox{90}{\textbf{Rec.}} & \rotatebox{90}{\textbf{F1}} & \rotatebox{90}{\textbf{Kappa}} & \rotatebox{90}{\textbf{MCC}} & \rotatebox{90}{\textbf{K-fold}} & \rotatebox{90}{\textbf{XAI}} &  \\ 
\hline
\cite{Akter2018} & \xmark & \xmark & \xmark & \cmark & \xmark & 8 & Supervised Learning & \cmark & \cmark & \cmark & \cmark & \xmark & \xmark & \xmark & \xmark & Low-precision model, smaller and simple dataset\\
\hline
\cite{Sarker2019} & \xmark & \xmark & \cmark & \cmark & \xmark & 8 & ORB & \cmark & \cmark & \cmark & \xmark & \xmark & \xmark & \xmark & \xmark & Model tested on smaller and simple dataset \\
\hline
\cite{12linkon2020deep} & \cmark & \cmark & \cmark & \cmark & \cmark & Multi & MobileNet, NasNetMobile, ResNet152 & \cmark & \cmark & \cmark & \xmark & \xmark & \xmark & \xmark & \xmark & Model provides inaccurate performance in noisy images\\
\hline
\cite{13joshi2020yolov3} & \xmark & \cmark & \xmark & \cmark & \cmark & 7 & YOLOv3 & \cmark & \cmark & \cmark & \cmark & \xmark & \xmark & \xmark & \xmark & Model tested on small dataset and has optimization issues\\
\hline
\cite{7tasnim2021bangladeshi} & \xmark & \xmark & \cmark & \cmark & \xmark & 8 & CNN based model & \cmark & \cmark & \cmark & \cmark & \xmark & \xmark & \xmark & \xmark & Model provides inaccurate performance in noisy image \\
\hline
\cite{PasumarthyYOLOV52022} & \xmark & \xmark & \cmark & \cmark & \cmark & 9 & YOLOV5 & \cmark & \cmark & \cmark & \xmark & \xmark & \xmark & \xmark & \xmark & Model tested on limited dataset; scalability concerns \\
\hline
\cite{8singh2022ipcrf} & \cmark & \xmark & \cmark & \cmark & \cmark & Multi & IPCRNet & \cmark & \cmark & \cmark & \cmark & \xmark & \xmark & \xmark & \cmark & Model effectiveness limited by dataset diversity, noisy images, and specificity to Indian currency \\
\hline
\cite{park2023mbdm} & \xmark & \cmark & \xmark & \cmark & \cmark & Multi & YOLO-V3, CNN & \cmark & \cmark & \cmark & \cmark & \xmark & \xmark & \cmark & \cmark & Poor performance with complex background and folded banknotes \\
\hline
\cite{Nasir2024} & \cmark & \xmark & \cmark & \cmark & \cmark & 9 & VGG19 & \cmark & \cmark & \cmark & \cmark & \xmark & \xmark & \xmark & \xmark & Limited to Pakistani Currency and limited evaluation \\
\hline
\cite{14IslamViT} & \xmark & \xmark & \cmark & \cmark & \xmark & 9 & ViT & \cmark & \cmark & \cmark & \cmark & \xmark & \xmark & \xmark & \cmark & Limited to Bangladeshi Currency and limited evaluation  \\
\hline
\cite{Siddiki2023}& \xmark & \xmark & \cmark & \cmark & \xmark & 9 & CNN & \cmark & \cmark & \cmark & \xmark & \xmark & \xmark & \xmark & \xmark & Very small test set (50 samples)\\
\hline
\cite{Chowdhury2024} & \cmark & \xmark & \cmark & \cmark & \cmark & 9 & VGG16, VGG19, InceptionV3, ResNet-V2 & \cmark & \cmark & \cmark & \cmark & \xmark & \xmark & \xmark & \xmark & Computational cost concerns, cross-dataset challenges \\
\hline
\cite{Das2023CNN} & \xmark & \xmark & \cmark & \cmark & \xmark & 9 & CNN & \cmark & \cmark & \cmark & \cmark & \xmark & \xmark & \xmark & \xmark & Insufficient real-world condition evaluation \\
\hline
\cite{Ali2024Egyptian} & \xmark & \xmark & \cmark & \cmark & \cmark & Multi & YOLOv10 & \cmark & \cmark & \cmark & \cmark & \xmark & \xmark & \xmark & \xmark & Small dataset (2,000), unspecified environmental diversity \\
\hline
\cite{Evwiekpaefe2024} & \xmark & \xmark & \cmark & \cmark & \cmark & 8 & MobileNetV2 & \cmark & \cmark & \cmark & \cmark & \xmark & \xmark & \xmark & \xmark & Controlled conditions, limited real-world evaluation \\
\hline
\cite{Neto2023Brazilian} & \xmark & \xmark & \cmark & \cmark & \xmark & Multi & CNN & \cmark & \cmark & \cmark & \xmark & \xmark & \xmark & \xmark & \xmark & Limited to Brazilian currency \\
\hline
\cite{Awad2022Iraqi} & \xmark & \xmark & \cmark & \cmark & \cmark & Multi & Deep CNN & \cmark & \cmark & \cmark & \cmark & \xmark & \xmark & \xmark & \xmark & Limited to Iraqi banknotes \\
\hline
\cite{Bolanos2024Colombian} & \xmark & \xmark & \cmark & \cmark & \xmark & Multi & Mobile App & \cmark & \cmark & \cmark & \xmark & \xmark & \xmark & \xmark & \xmark & Limited to Colombian currency \\
\hline
\textbf{Proposed Hybrid Model} & \cmark & \cmark & \cmark & \cmark & \cmark & 9 & MobileNetV3, EfficientNetB0, MLP & \cmark & \cmark & \cmark & \cmark & \cmark & \cmark & \cmark & \cmark & This framework is confined within two stages, feature extraction and classification, and is limited to Bangladeshi \\
\hline
\end{tabular}
}
\vspace{0.5cm}
\small
\\ \textbf{Abbreviations:} Acc=Accuracy, Pre=Precision, Rec=Recall, Kappa=Cohen's Kappa, MCC=Matthews Correlation Coefficient, K-fold=K-fold Cross-Validation, XAI=Explainable AI, ORB=Oriented FAST and Rotated BRIEF, CNN=Convolutional Neural Network, YOLO=You Only Look Once, ViT=Vision Transformer, MLP=Multilayer Perceptron, \cmark=Yes, \xmark=No
\end{table*}

Significant research has been conducted in the past few years on currency recognition for visually impaired individuals, with a wide range of machine learning and deep learning algorithms employed. Despite significant progress in currency recognition, notable limitations persist. First, most existing systems rely on a single-architecture model, thereby missing opportunities for hybrid models that integrate complementary feature-extraction strengths. Second, dataset diversity is still inadequate; many studies evaluate performance under homogeneous or controlled conditions, without systematically evaluating robustness across progressive complexity, such as varying illumination, occlusions, cluttered backgrounds, and diverse viewing angles. For instance, \cite{Ali2024Egyptian} and Nigerian \cite{Evwiekpaefe2024} systems used relatively smaller datasets with limited diversity, while Indian studies \cite{13joshi2020yolov3}, \cite{8singh2022ipcrf},\cite{PasumarthyYOLOV52022} employed larger datasets but provided minimal documentation of data acquisition conditions. Similarly, the reviewed studies on Bangladeshi currency datasets \cite{12linkon2020deep},\cite{7tasnim2021bangladeshi},\cite{BanglaTaka2025},\cite{Siddiki2023},\cite{Chowdhury2024},\cite{Das2023CNN},\cite{14IslamViT} did not present proper systematic evaluation across the spectrum of complexity. Although the BanglaNotes dataset \cite{BanglaTaka2025} offers high-quality images, it only contains 5,073 images and does not incorporate progressive complexity evaluation. Third, despite its significance in ensuring debugging, transparency, and user confidence in assistive technologies, Explainable AI integration remains highly underutilized. Fourth,  optimization methodology is a rarely focused area in which most fundamental analysis work is conducted using one optimizer with predefined learning rates with a basic split for model evaluation, not utilizing k-fold cross-validation with a rigorous evaluation of statistical analysis. Currency recognition for Bangladeshi banknotes has evolved from early supervised learning methodologies ~\cite{Akter2018} through convolutional neural network architectures ~\cite{7tasnim2021bangladeshi}, \cite{Siddiki2023} to precision-oriented identification systems ~\cite{Das2023CNN}, transfer learning models ~\cite{Chowdhury2024}, and recently Vision Transformers ~\cite{14IslamViT}. While these works indicate increasing architectural complexity, they lack systematic real-world evaluation, hybrid model designs, optimizer analysis, cross-validation methods, and model explainability mechanisms.

To address these gaps, a novel hybrid architecture has been proposed in this study that combines MobileNetV3-Large and EfficientNetB0, followed by an MLP classifier. We have built five progressively complex datasets totaling 8,100 images spanning controlled to real-world conditions to enable robust performance evaluation. We further conduct an extensive analysis of four optimizers across multiple learning rates to identify effective training strategies. Model reliability was evaluated through five-fold cross-validation using seven performance metrics, including Accuracy, Precision, Recall, F1-Score, Cohen's Kappa, MCC, and AUC. Additionally, we integrate LIME and SHAP to provide interpretable explanations for transparent insights into model decision-making. A comparative analysis of the related works in currency recognition systems has been presented in Table~\ref{tab:related}.

\section{Methodology}
\label{section:methodology}
\begin{figure*}[t] 
    \centering
    \includegraphics[width=\linewidth]{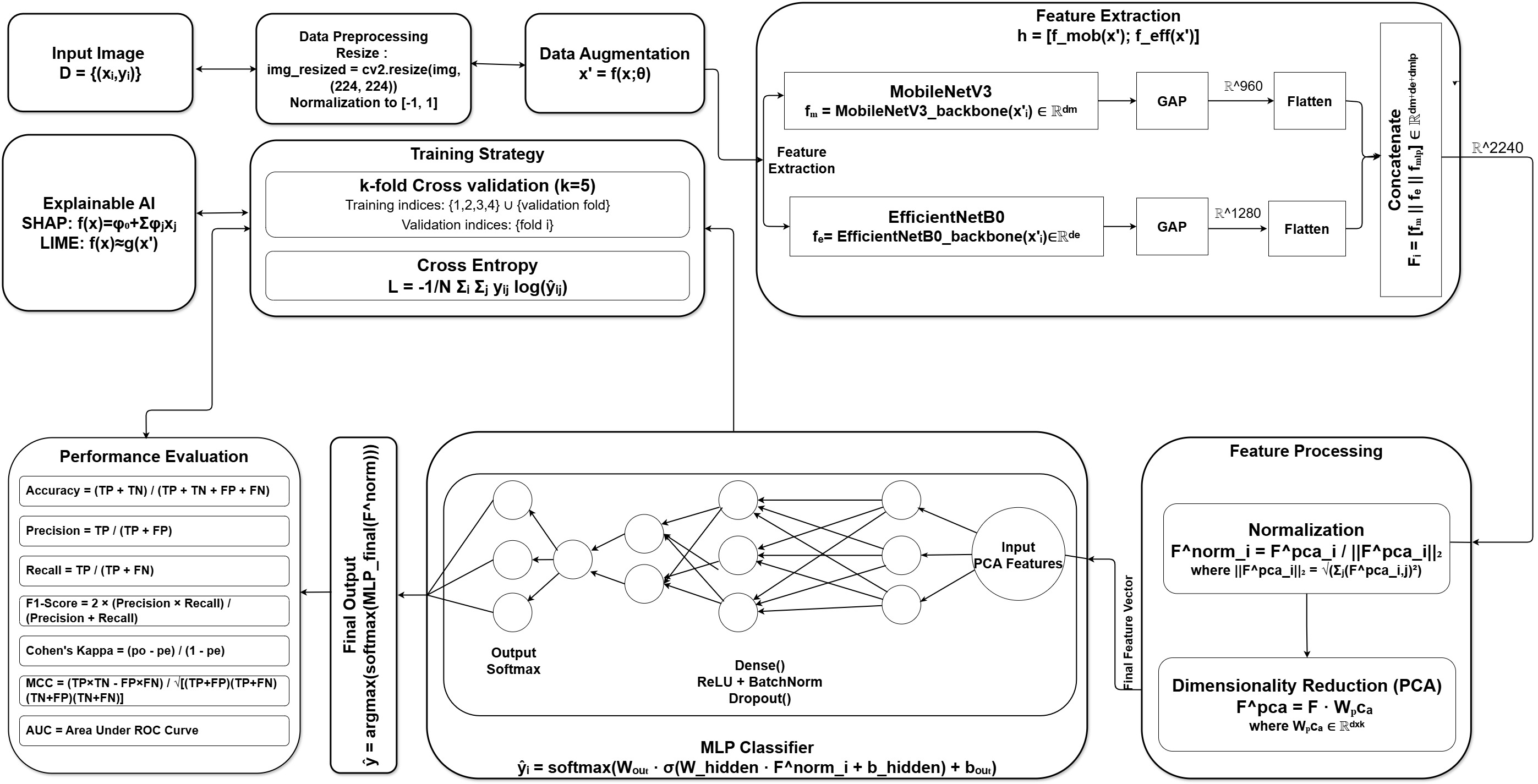}
    \caption{Workflow of the Proposed Hybrid Model for Currency Recognition}
    \label{fig:workflow}
\end{figure*}

In this section, we have presented the methodology of our research. This study has aimed to develop a novel hybrid model for currency detection to aid assistive technologies for currency recognition. We began by explaining the procedure for building the dataset. Next, we present the architecture of the proposed hybrid model, comprising a feature extraction stage and a classification stage. We then elaborated on the methodology's experimental settings. Finally, we have concluded by interpreting the evaluation metrics used for the proposed model. The overall workflow of the proposed hybrid model for currency recognition is illustrated in Figure~\ref{fig:workflow}.

\subsection{Dataset Construction}
In this section, we have provided a detailed description of the dataset used. Building a dataset involves data collection, preprocessing, augmentation, defining the dataset structure, splitting, and exporting it to train the hybrid model.

\subsubsection{Data Collection}
\begin{figure*}[htbp]

  \centering
  
  \begin{subfigure}[b]{0.22\textwidth}
    \includegraphics[width=\textwidth]{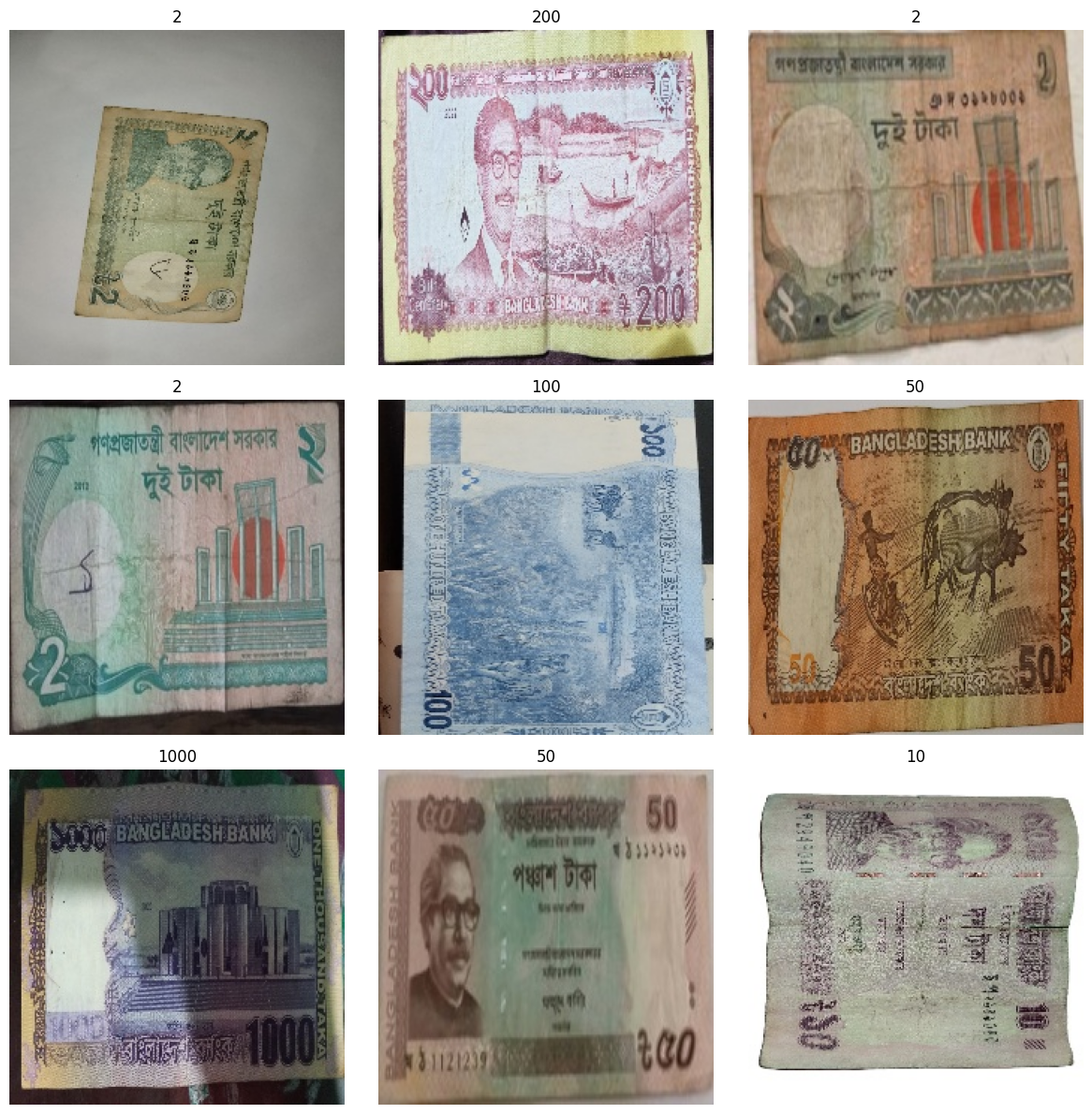}
    \caption{Primary Dataset-1}
  \end{subfigure}
  \hfill
  \begin{subfigure}[b]{0.22\textwidth}
    \includegraphics[width=\textwidth]{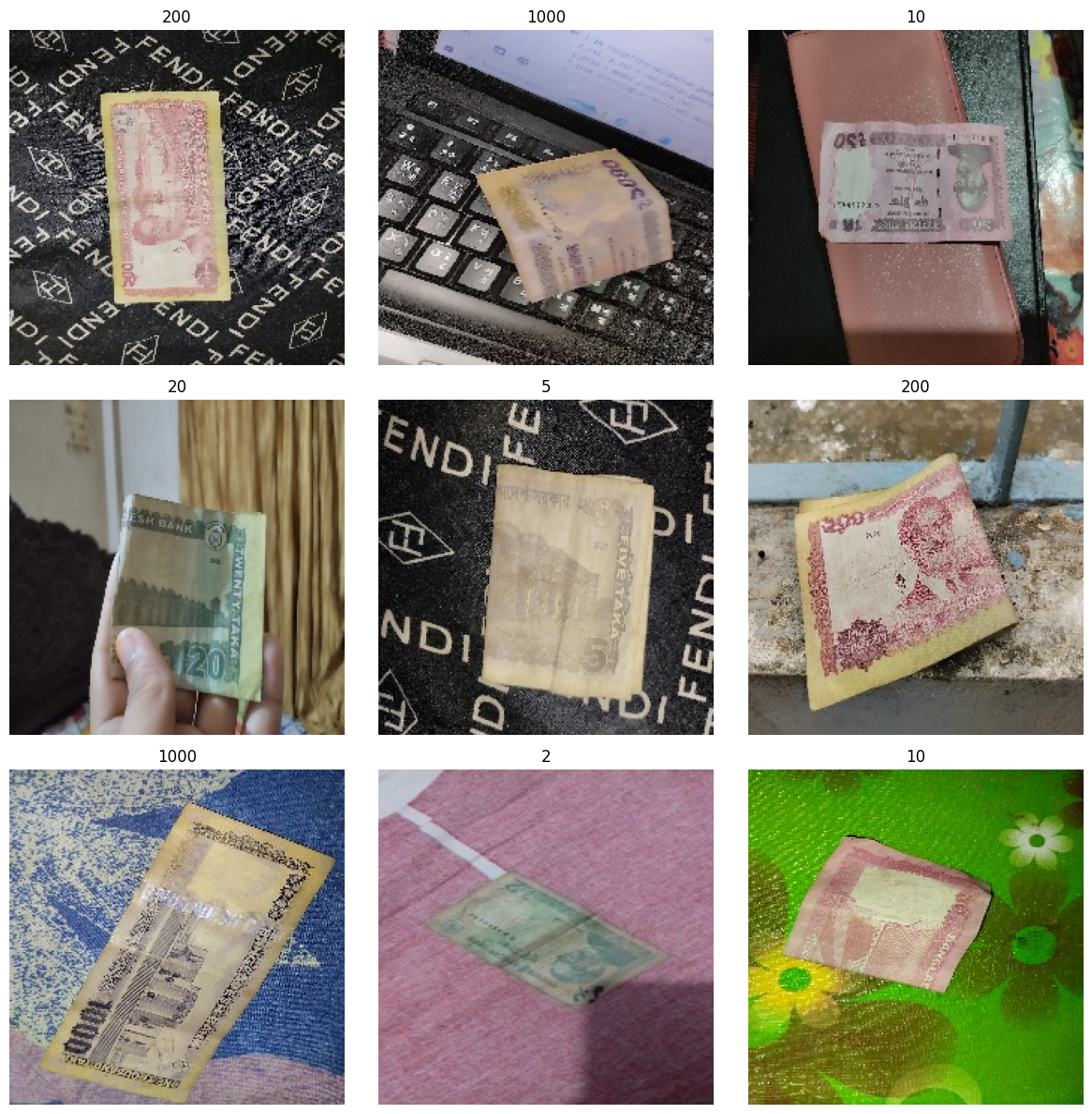}
    \caption{Primary Dataset-2}
  \end{subfigure}
  \hfill
  \begin{subfigure}[b]{0.22\textwidth}
    \includegraphics[width=\textwidth]{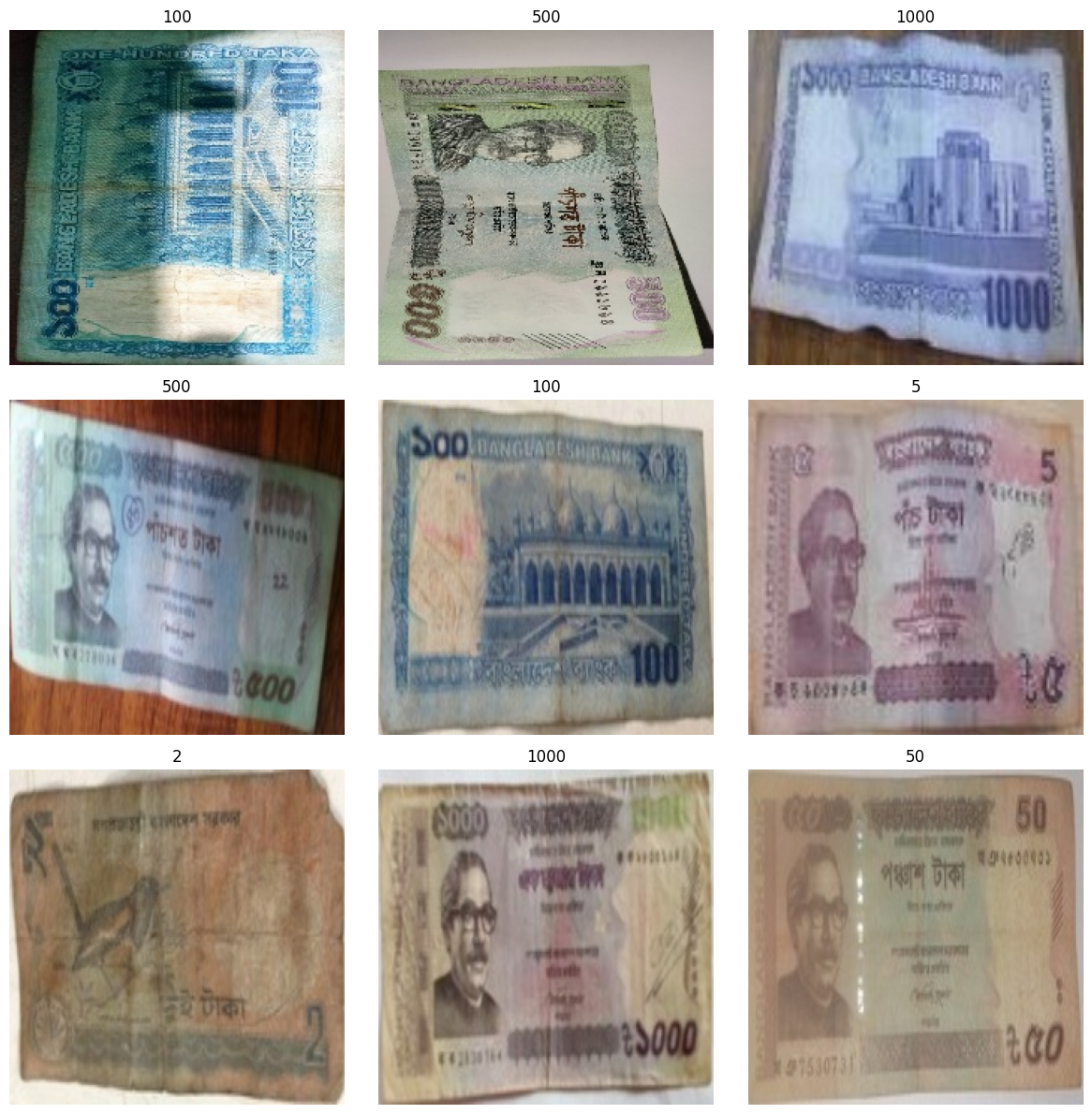}
    \caption{Custom Dataset-2}
  \end{subfigure}
  \hfill
  \begin{subfigure}[b]{0.22\textwidth}
    \includegraphics[width=\textwidth]{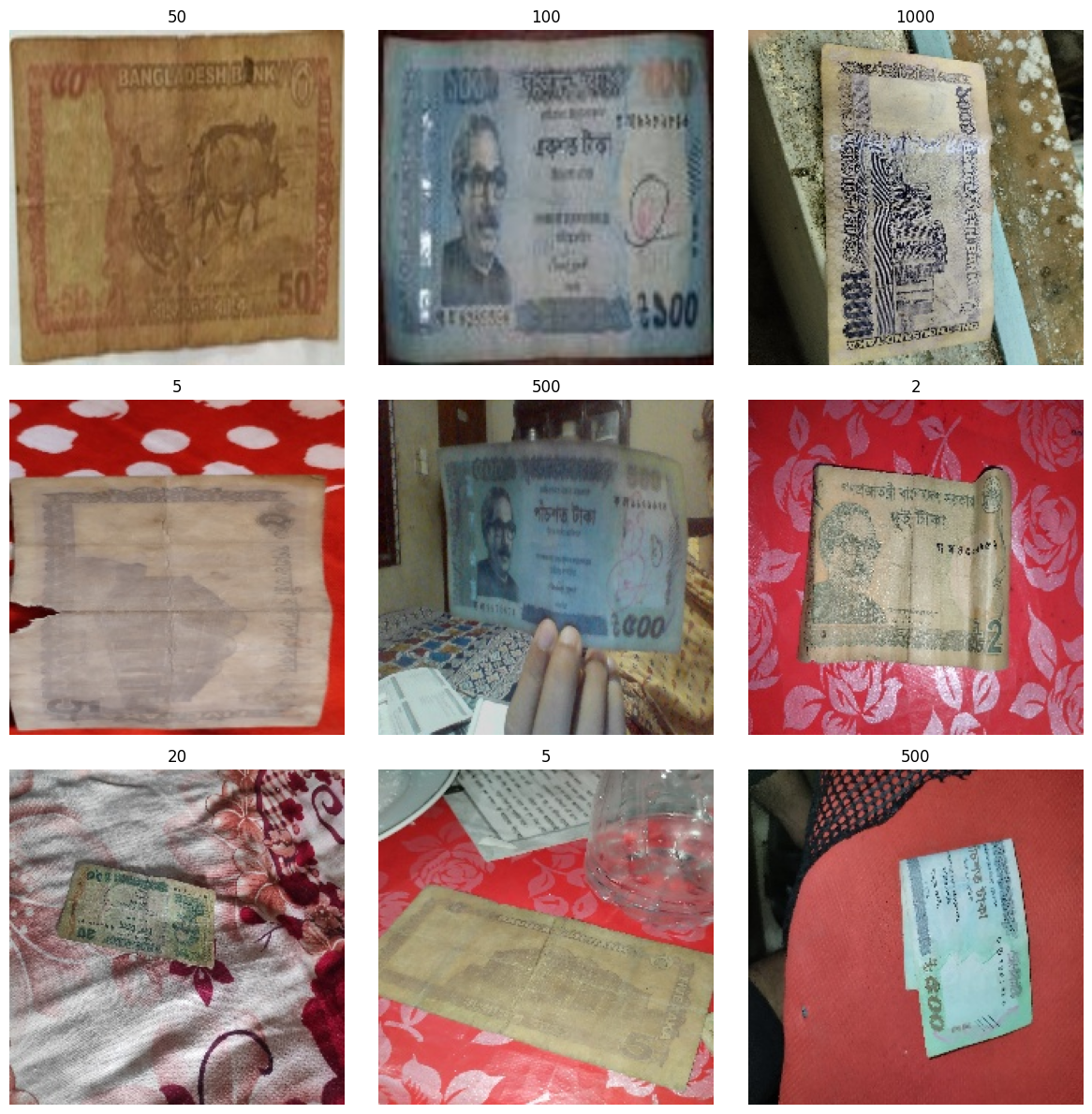}
    \caption{Custom Dataset-3}
  \end{subfigure}
  \caption{Representative samples from the datasets highlighting the progressive background complexity}
 \label{fig:dataset_sample}
\end{figure*}

Data collection is the first and fundamental step of research. To build our Bangladeshi banknote datasets, we have collected images of the currently available and frequently used Bangladeshi currency in 9 denominations (2 Tk, 5 Tk, 10 Tk, 20 Tk, 50 Tk, 100 Tk, 200 Tk, 500 Tk, and 1000 Tk). The images were collected by taking pictures from a mobile camera, a laptop webcam, and from publicly available datasets \cite{Banglacurrency2019}, \cite{BanglaMoney2018}, \cite{AugmentedBanglaMoney}. Figure~\ref{fig:dataset_sample} shows representative samples from the datasets that illustrate the progressive increase in visual variability and environmental complexity. 
Formally, the complete dataset $D$ can be expressed as the union of the sources:
\begin{equation}
    D = D_{\text{mobile}} \cup D_{\text{webcam}} \cup D_{\text{public}},
\end{equation}
where $D_{\text{mobile}}$, $D_{\text{webcam}}$, and $D_{\text{public}}$ represent the sets of images collected using mobile devices, laptop webcam, and publicly available datasets, respectively.  \\
Moreover, considering the nine denominations, the dataset can also be represented as a summation of images across all denominations and sources:
\begin{equation}
    D = \sum_{i=1}^{9} \left( d_{i}^{\text{mobile}} + d_{i}^{\text{webcam}} + d_{i}^{\text{public}} \right),
\end{equation}
where $d_{i}^{\text{mobile}}$, $d_{i}^{\text{webcam}}$, and $d_{i}^{\text{public}}$ denote the number of images of the $i$-th denomination obtained from mobile camera, laptop webcam, and public datasets, respectively. 
\subsubsection{Dataset Description}
A key contribution of this work is the curation of five Bangladeshi banknote datasets that represent progressively increasing environmental complexity.  The composition and characteristics of each dataset are summarized in Table~\ref{tab:dataset_overview}. 


\begin{table*}[htbp]
\centering
\caption{Overview and Partitioning of Bangladeshi Banknote Datasets}
\label{tab:dataset_overview}
\begin{tabular}{lccccl}
\toprule
\textbf{Dataset} & \textbf{Total} & \textbf{Train} & \textbf{Val.} & \textbf{Test} & \textbf{Characteristics} \\
\midrule
Custom Dataset-1 & 2,700 & 1,836 & 459 & 405 & Public benchmarks with controlled background \\

Primary Dataset-1 & 2,250 & 1,530 & 387 & 342 & Simple backgrounds, consistent illumination \\
Custom Dataset-2 & 4,950 & 3,366 & 837 & 747 & Combined Primary Dataset-1 + Primary Dataset-2 \\
Primary Dataset-2 & 2,700 & 1,836 & 459 & 405 & Complex backgrounds, varied lighting, occlusions \\
Custom Dataset-3 & 8,100 & 5,508 & 1,377 & 1,215 & All datasets combined, maximum diversity \\
\bottomrule
\end{tabular}
\vspace{0.5em}

\small
\textit{Note:} Dataset progression ranges from controlled laboratory conditions to complex real-world scenarios, with class distribution balanced across all denominations with minor variations ($\pm$5\%)  to reflect real-world currency circulation patterns for model evaluation.  
\end{table*}

\textbf{Custom dataset-1},Sourced from publicly available repositories \cite{BanglaMoney2018}, \cite{Banglacurrency2019}, and \cite{AugmentedBanglaMoney}, this baseline dataset contains 2,700 images captured under controlled conditions with uniform white backgrounds and standardized lighting. This dataset serves as the foundation for initial model training and establishes an upper performance bound under ideal conditions.

\textbf{Primary dataset-1},  Captured using a smartphone camera in environments with simple, uncluttered backgrounds or plain surfaces. This dataset, comprising 2,250 images, represents an intermediate complexity level with controlled but naturally varying background conditions.

\textbf{Custom Dataset-2} A hybrid dataset combining the Dataset-1 and Primary Dataset- (4,950 images total), tailored to assess model performance when trained on mixed-complexity data. 

\textbf{Primary dataset-2 }Captured in real-world scenarios, including environments with complex backgrounds, varying illumination (indoor/outdoor, natural/artificial lighting), partial occlusions (hands, objects), and diverse viewing angles. This dataset (2,700 images) represents the most challenging evaluation scenario closest to practical deployment conditions.

\textbf{Custom Dataset-3} The comprehensive dataset aggregating all previous datasets plus additional augmented samples, totaling 8,100 images. This maximum-diversity dataset enables evaluation of model generalization across the full spectrum of acquisition conditions.

\subsection{Model Architecture}
\label{sec:architecture}

\begin{figure*}[t] 
    \centering
    \includegraphics[width=\linewidth]{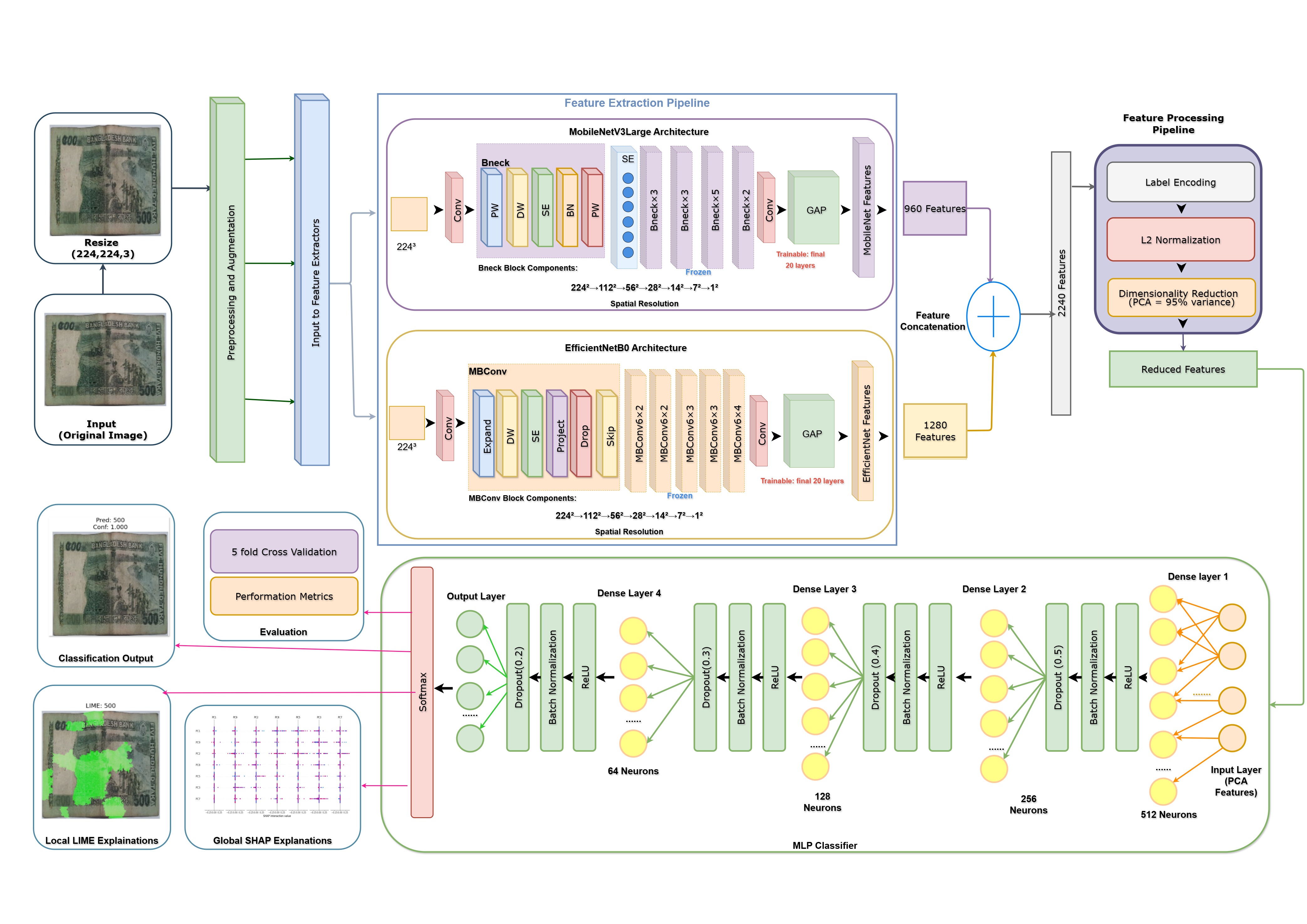}
    \caption{Model Architecture of the Proposed Hybrid Currency Recognition Model}
    \label{fig:architecture}
\end{figure*}

The proposed system utilized a hybrid deep learning architecture, integrating two pretrained convolutional neural networks, MobileNetv3-Large and EfficientNetB0, followed by a compact multilayer perceptron (MLP) classifier. The selection of architecture was effective since it balanced achieving high recognition accuracy and was effective in terms of computational performance, making it suitable to implement on resource-limited devices. The pipeline comprised three steps: (i) a dual-branch feature extraction pipeline using transfer learning, (ii) a feature processing pipeline, and (iii) classification by MLP. Figure~\ref{fig:architecture} illustrates the complete architectural pipeline of the proposed system. Algorithm~\ref{alg:hybrid} summarizes the overall pipeline, including the steps for feature extraction, feature fusion, normalization, dimensionality reduction, classification, and explainability.

\begin{algorithm}
\caption{Hybrid Model for Banknote Classification with Extended Metrics}
\label{alg:hybrid}
\begin{algorithmic}[1]
\REQUIRE Training set $\mathcal{D}_{train} = \{(\mathbf{x}_i, y_i)\}_{i=1}^{N}$, Test set $\mathcal{D}_{test} = \{(\mathbf{x}_j, y_j)\}_{j=1}^{M}$ where $\mathbf{x} \in \mathbb{R}^{224 \times 224 \times 3}$
\ENSURE Final Model $\mathcal{M}^*$, Performance metrics $\Psi$, Explainability analysis $\{\mathcal{E}_{LIME}, \boldsymbol{\Phi}_{SHAP}\}$

\STATE Initialize random seed and augmentation: $\theta_{aug}=\{\pm30^\circ,\pm10\%,[0.8,1.2]\}$
\STATE Load pre-trained CNNs: $\mathcal{F}_{mob} \leftarrow$ MobileNetV3Large, $\mathcal{F}_{eff} \leftarrow$ EfficientNetB0
\STATE Freeze all layers except final 20: $\mathcal{F}_{mob}[-20:], \mathcal{F}_{eff}[-20:] \leftarrow$ Trainable

\FOR{$\mathbf{x}_i \in \mathcal{D}_{train}$}
    \STATE Generate augmented image $\mathbf{x}_i^{aug} \leftarrow \mathcal{A}(\mathbf{x}_i)$
    \STATE Extract MobileNetV3 features: $\mathbf{f}_{mob} \leftarrow \text{GAP}(\mathcal{F}_{mob}(\mathbf{x}_i))$
    \STATE Extract EfficientNetB0 features: $\mathbf{f}_{eff} \leftarrow \text{GAP}(\mathcal{F}_{eff}(\mathbf{x}_i))$
    \STATE Concatenate: $\mathbf{f}_i^{orig} \leftarrow [\mathbf{f}_{mob} \oplus \mathbf{f}_{eff}] \in \mathbb{R}^{2240}$
    \STATE Repeat for augmented: $\mathbf{f}_i^{aug} \leftarrow [\text{GAP}(\mathcal{F}_{mob}(\mathbf{x}_i^{aug})) \oplus \text{GAP}(\mathcal{F}_{eff}(\mathbf{x}_i^{aug}))]$
    \STATE Store: $\mathcal{F}_{train} \leftarrow \mathcal{F}_{train} \cup \{\mathbf{f}_i^{orig}, \mathbf{f}_i^{aug}\}$, $\mathcal{Y}_{train} \leftarrow \mathcal{Y}_{train} \cup \{y_i, y_i\}$
\ENDFOR

\STATE Apply L2 normalization: $\hat{\mathbf{f}}_i \leftarrow \mathbf{f}_i / \|\mathbf{f}_i\|_2$ $\forall \mathbf{f}_i \in \mathcal{F}_{train}$
\STATE PCA: $\mathbf{Z} = \hat{\mathbf{F}} \mathbf{V}_K$ where $\sum_{k=1}^{K} \lambda_k / \sum_{j=1}^{2240} \lambda_j \geq 0.95$

\FOR{$k = 1$ to $5$} 
    \STATE Define MLP: 
    $\mathcal{M}_k \leftarrow \text{Dense}(256) \to \text{ReLU+BN+DO}(0.5) \to$
    \STATE \hspace{3.5em} $\text{Dense}(128) \to \text{ReLU+BN+DO}(0.4) \to \text{Dense}(64) \to$
    \STATE \hspace{3.5em} $\text{ReLU+BN+DO}(0.3) \to \text{Softmax}(C)$
    \STATE Split: $(\mathbf{Z}_{train}^{(k)}, \mathbf{Z}_{val}^{(k)}) \leftarrow \text{StratifiedSplit}(\mathbf{Z}, \mathcal{Y}_{train}, k)$
    \STATE Train: $\mathcal{M}_k^* \leftarrow \arg\min_{\boldsymbol{\theta}} \mathbb{E}[\mathcal{L}_{CCE}(\mathcal{M}_k(\mathbf{Z}_{tr}^{(k)}), \mathbf{y}_{tr}^{(k)})]$
    \STATE Evaluate: $\psi_k \leftarrow \{\text{Acc}, \text{Prec}, \text{Rec}, \text{F}_1, \text{Kappa}, \text{MCC}, \text{AUC}\}$
    \STATE \hspace{3.5em} on $(\mathbf{y}_{val}^{(k)}, \arg\max \mathcal{M}_k^*(\mathbf{Z}_{val}^{(k)}))$
\ENDFOR

\STATE Select best: $k^* \leftarrow \arg\max_k \psi_k[\text{Acc}]$
\STATE Retrain final: $\mathcal{M}^* \leftarrow \text{Retrain}(\mathbf{Z}, \mathcal{Y}_{train}; \text{config}_{k^*})$

\STATE Extract test features: $\mathbf{Z}_{test} \leftarrow \mathbf{V}_K^T \cdot \text{L2Norm}([\mathcal{F}_{mob}(\mathcal{D}_{test}) \oplus$
\STATE \hspace{3.5em} $\mathcal{F}_{eff}(\mathcal{D}_{test})])$
\STATE Generate predictions: $\hat{\mathbf{Y}}_{test} \leftarrow \arg\max \mathcal{M}^*(\mathbf{Z}_{test})$
\STATE Test metrics: $\Psi \leftarrow \{\text{Accuracy}, \text{Precision}, \text{Recall}, \text{F}_1,$
\STATE \hspace{3.5em} $\text{CohenKappa}, \text{MCC}, \text{AUC}\}$
\STATE Generate ROC: $\text{ROC} \leftarrow \text{ROC\_Curve}(\mathbf{Y}_{test}, \hat{\mathbf{Y}}_{test})$

\STATE \textbf{LIME:} Predictor $\Pi_{LIME}(\mathbf{x})=\mathcal{M}^*(P(\text{L2Norm}([\mathcal{F}_{mob}(\mathbf{x})\oplus\mathcal{F}_{eff}(\mathbf{x})])))$
\STATE Generate LIME explanations $\mathcal{E}_{LIME}$
\FOR{$s \in \mathcal{S}_{LIME}$}
    \STATE $\mathcal{E}_{LIME}^{(s)} \leftarrow \text{LIMEExplainer}(\mathbf{x}_s, \Pi_{LIME}; n_{samples}=1000)$
\ENDFOR

\STATE \textbf{SHAP:} $\mathcal{B} \leftarrow \text{RandomSample}(\mathbf{Z}, 50)$
\STATE Global SHAP: $\Phi_{exp} \leftarrow \text{KernelExplainer}(\mathcal{M}^*, \mathcal{B})$
\STATE $\boldsymbol{\Phi} \leftarrow \Phi_{exp}.\text{shap\_values}(\text{RandomSample}(\mathbf{Z}_{test}, 100); n_{samples}=200)$

\STATE Aggregate: $\mu_{CV} \pm \sigma_{CV} \leftarrow \text{Stats}(\{\psi_k\}_{k=1}^5)$

\RETURN $\mathcal{M}^*$, $\{\mu_{CV} \pm \sigma_{CV}, \Psi, \text{ROC}\}$, $\{\mathcal{E}_{LIME}, \boldsymbol{\Phi}_{SHAP}\}$

\end{algorithmic}
\end{algorithm}
\subsection{Dual-Branch Feature Extraction}

A parallel dual-branch feature-extraction architecture was employed, using MobileNetV3-Large and EfficientNetB0, both pretrained on ImageNet. MobileNetV3-Large is an efficiency-optimized CNN discovered via neural architecture search (NAS), designed for low-resource devices \cite{ Howard2019mobv3}, whereas EfficientNetB0 uses compound scaling of depth, width, and resolution with Mobile Inverted Bottleneck (MBConv) blocks \cite{tan2019efficientnet}. The classification heads are removed to preserve the convolutional feature extractors.

\subsection{Feature Fusion Pipeline}

Feature vectors obtained from both branches, $\mathbf{f}_{\text{MobileNet}}, \mathbf{f}_{\text{EfficientNet}} \in \mathbb{R}^{1280}$, are concatenated to form a unified representation:
\begin{equation}
    \mathbf{f}_{\text{concat}} = [\mathbf{f}_{\text{MobileNet}}; \mathbf{f}_{\text{EfficientNet}}] \in \mathbb{R}^{2560}
\end{equation}

This dual-branch configuration leveraged complementary features, where MobileNetV3 captures efficient edge and texture patterns, and EfficientNetB0 provides multi-scale representations to enhance robustness for subsequent MLP classification.

\subsection{Feature Processing Pipeline}

Original RGB images were processed by MobileNetV3-Large and EfficientNetB0 architectures for feature extraction.  The classification heads were stripped, the final 20 layers were trained, and global average pooling produced compact feature embeddings. The embeddings were then added together to create a single feature representation. 

\subsubsection{L2 Normalization}
The concatenated feature vector $\mathbf{f}_{\text{concat}} \in \mathbb{R}^{2560}$ were L2-normalized:
\begin{equation}
    \mathbf{f}' = \frac{\mathbf{f}_{\text{concat}}}{\|\mathbf{f}_{\text{concat}}\|_2}
\end{equation}
This ensures scale invariance, mitigates magnitude differences between CNNs, emphasizes angular relationships, stabilizes gradients, and improves numerical stability. 

\subsubsection{Principal Component Analysis (PCA)}

PCA reduced dimensionality while preserving 95\% of variance. Given normalized features $\mathbf{X} \in \mathbb{R}^{N \times 2560}$:
\begin{align*}
    \mathbf{C} &= \frac{1}{N}\mathbf{X}^T\mathbf{X}, \quad
    \mathbf{C} = \mathbf{V}\mathbf{\Lambda}\mathbf{V}^T, \\
    \mathbf{X}_{\text{PCA}} &= \mathbf{X} \cdot \mathbf{V}_k
\end{align*}
This reduces features from 2560 to $k \approx 200$--400 while preserving discriminative information and lowering computational cost.

\subsection{Compact Multilayer Perceptron Classifier}

The reduced feature vector was input to an MLP classifier with fully connected layers forming a bottleneck architecture. Each hidden layer uses ReLU activation, batch normalization, and progressive dropout ($0.5 \rightarrow 0.2$). The final softmax layer outputs class probabilities:
\begin{equation}
    P(y=i|\mathbf{x}) = \frac{\exp(z_i)}{\sum_{j=1}^{C} \exp(z_j)}, \quad i=1,\ldots,C
\end{equation}

Table~\ref{tab:mlp_architecture} summarizes the layer-wise architecture of MLP used in this study:


It is a lightweight architecture ($\sim$327K parameters) that employs fewer parameters than most models, but with high classification accuracy, and is thus suitable to run on resource-constrained devices in real-time. The concept of batch normalization balances the training process, and the concept of progressive dropout decreases the chances of overfitting.

\subsection{Evaluation Metrics}
Multiple evaluation metrics were employed to evaluate the performance of the proposed hybrid model. These metrics not only assess the classification accuracy but also robustness, generalization, and interpretability. 

\subsubsection{Accuracy}
Accuracy is the fundamental evaluation metric to assess the performance of a model. It is calculated by the formula: 
\begin{equation}
\text{Accuracy} = \frac{TP + TN}{TP + TN + FP + FN},
\end{equation}
where $TP$, $TN$, $FP$, and $FN$ denote true positives, true negatives, false positives, and false negatives, respectively.  

\subsubsection{Precision}
Precision is the indication of the proportion of true positives among all predicted positives made by the model. Mathematically, it is expressed by:
\begin{equation} 
    \text{Precision} = \frac{TP}{TP+FP}
\end{equation} 

\subsubsection{Recall}
Also known as sensitivity rate or True Positive Rate (TPR), recall is the indication of the proportion of true positives among all the actual positives. It is calculated by:
\begin{equation}
\text{Recall} = \frac{TP}{TP + FN}.
\end{equation} 

\subsubsection{F1-score}
The F1-score is the harmonic mean of precision and recall. It is expressed mathematically as:
\begin{equation}
F1 = 2 \times \frac{\text{Precision} \times \text{Recall}}{\text{Precision} + \text{Recall}}.
\end{equation}

\subsubsection{Cohen Kappa Coefficient}
Cohen's Kappa Coefficient a is a statistic that measures inter-rater reliability of qualitative items by comparing observed accuracy with expected accuracy from random chance. It is calculated by: 
\begin{equation}
\kappa = \frac{p_o - p_e}{1 - p_e},
\end{equation}
where $p_o$ is the observed agreement among raters and $p_e$ is the hypothetical probability of chance agreement, which is calculated using observed data. A value of $\kappa$ > 0.80 is indicated as strong agreement. 

\subsubsection{Matthews Correlation Coefficient}
Matthews Correlation Coefficient (MCC) is a metric that is useful in the case of imbalanced data, where traditional evaluation metrics can be misleading. It is used to evaluate all four confusion matrix categories. It is expressed mathematically as:
\begin{equation}
\text{MCC} = \frac{TP \times TN - FP \times FN}{\sqrt{(TP + FP)(TP + FN)(TN + FP)(TN + FN)}}.
\end{equation}
MCC values ranging from $-1$ (total disagreement) to $+1$ (perfect agreement). 

\subsubsection{ROC curve and AUC score}
ROC (Receiver Operator Characteristic) curve is a graphical representation that illustrates the tradeoff between true positive rate and false positive rate. On the other hand, AUC (Area Under Curve) quantifies the separability of classes by calculating the area under the ROC curve. 
\begin{equation}
\text{AUC} = \int_0^1 \text{TPR}(x) \, d(\text{FPR}(x)).
\end{equation}
The value of AUC ranges from 0 to 1, where a value close to 1 indicates strong discriminative capability.
\subsubsection{Cross-validation}
Cross-validation is a statistical method to assess the generalization capability of a model and mitigate overfitting.
\begin{equation}
\text{CV\_Score} = \frac{1}{k} \sum_{i=1}^{k} M_i,
\end{equation}
where $M_i$ denotes the metric from the $i$-th fold. 

\subsubsection{Explainable AI methods}
Explainable AI (XAI) methods provide an understanding of the decision-making process by highlighting which features or regions of the input contribute most to its predictions. To enhance transparency and interpretability, two Explainable AI (XAI) methods were employed: LIME and SHAP. 

\textbf{LIME (Local Interpretable Model-Agnostic Explanations):} It is an XAI approach that explains individual predictions by approximating the complex model locally with an interpretable surrogate model $g \in G$, where $G$ is the class of simple models. \\
\begin{equation}
\underset{g \in G}{\arg\min} \; L(f, g, \pi_x) + \Omega(g),
\end{equation}
where $L$ measures the fidelity between $f$ and $g$ in the locality defined by $\pi_x$, and $\Omega(g)$ penalizes the complexity of $g$.  \\
\textbf{SHAP (SHapley Additive exPlanations):} It is another XAI approach that interprets model predictions by assigning importance scores to input features. This XAI method is based on Shapley values from cooperative game theory, providing theoretically consistent feature attributions of the proposed model. 
\begin{equation}
\phi_i = \sum_{S \subseteq F \setminus \{i\}} \frac{|S|!(|F|-|S|-1)!}{|F|!} 
\big[ f(S \cup \{i\}) - f(S) \big],
\end{equation}
where $F$ is the set of all features and $S$ is a subset of features not including $i$.

\begin{table*}[htbp]
\centering
\caption{MLP Classifier Architecture Specifications}
\label{tab:mlp_architecture}

\begin{tabular}{lccccc}
\toprule
\textbf{Layer} & \textbf{Input} & \textbf{Output} & \textbf{Activation} & \textbf{Dropout} & \textbf{Parameters} \\
\midrule

Layer 1 & $k$ & 512 & ReLU & 0.5 & $(k+1)\times512$ \\
Layer 2 & 512 & 256 & ReLU & 0.4 & 131,328 \\
Layer 3 & 256 & 128 & ReLU & 0.3 & 32,896 \\
Layer 4 & 128 & 64 & ReLU & 0.2 & 8,256 \\
Layer 5 & 64 & $C$ & Softmax & -- & $65\times C$ \\

\midrule
\multicolumn{5}{r}{\textbf{Total Parameters}} & $\sim$327,665\textsuperscript{*} \\

\bottomrule
\end{tabular}
\end{table*}

\subsection{Experimental setting}
In the following, we have discussed the experimental settings and configuration to optimize the performance of the proposed model.

\subsubsection{Computing Resource}
All experiments were conducted on the cloud-based platforms Google Colab and Kaggle Notebook. Python~3.10, TensorFlow~2.x, and supporting libraries such as Scikit-learn, OpenCV, and LIME were included in the software stack. 

\subsubsection{Training Configuration}
The training and evaluation of the proposed hybrid model followed a stratified 5-fold cross-validation strategy to ensure rigor and minimize bias. For each fold, 80\% of the data were used for training and 20\% for validation, while data augmentation was applied dynamically during training. A batch size of 32 was configured across all experiments. Early stopping with a patience of 10 epochs and a learning rate scheduler (ReduceLROnPlateau) was employed to prevent overfitting.

\subsubsection{Learning Rate}
The model was optimized using an adaptive learning rate strategy. During the training process, different learning rates like $\eta = 10^{-1}$, $\eta = 10^{-2}$, $\eta = 10^{-3}$, and $\eta = 10^{-4}$ were utilized to obtain the best learning rate for the proposed model. Note that the initial learning rate was set and reduced by a factor of $0.1$ when the validation loss plateaued for five consecutive epochs. Formally:

\begin{equation}
\eta_{t+1} = 
\begin{cases}
0.1 \cdot \eta_t, & \text{if no improvement for 5 epochs}, \\
\eta_t, & \text{otherwise}.
\end{cases}
\end{equation}

\subsubsection{Loss Function}
The classification task involved predicting one of $C$ the banknote denominations. Hence, categorical cross-entropy loss was employed:
\begin{equation}
\mathcal{L}_{CCE} = - \sum_{i=1}^{N} \sum_{c=1}^{C} y_{i,c} \cdot \log(\hat{y}_{i,c}),
\end{equation}
where $N$ is the number of samples, $y_{i,c}$ is the ground-truth one-hot encoding for class $c$, and $\hat{y}_{i,c}$ is the predicted probability.

\subsubsection{Optimizer}

To identify the most suitable optimizer for our datasets, we evaluated four optimization algorithms: Adam, AdamW, SGD, and RMSProp. These optimizers were chosen for their proven efficiency in employing adaptive learning rates and handling sparse gradients, which are important for training deep neural networks.

\textbf{Adam Optimizer} or
The Adaptive Moment Estimation (Adam) algorithm combines the advantages of both AdaGrad and RMSProp. Its update rule for parameter $\theta$ at iteration $t$ is given by:

\begin{align}
m_t &= \beta_1 m_{t-1} + (1-\beta_1) g_t, \\
v_t &= \beta_2 v_{t-1} + (1-\beta_2) g_t^2, \\
\hat{m}_t &= \frac{m_t}{1-\beta_1^t}, \quad \hat{v}_t = \frac{v_t}{1-\beta_2^t}, \\
\theta_{t+1} &= \theta_t - \eta \frac{\hat{m}_t}{\sqrt{\hat{v}_t} + \epsilon},
\end{align}
where $g_t = \nabla_{\theta_t}\mathcal{L}(\theta_t)$ is the gradient of the loss function at iteration $t$, $m_t$ and $v_t$ represent the first and second moment estimates, $\beta_1=0.9$ and $\beta_2=0.999$ are the exponential decay rates, $\eta$ is the learning rate, and $\epsilon = 10^{-8}$ is a small constant for numerical stability.

\textbf{AdamW Optimizer}
decouples weight decay from the gradient-based update, addressing the issue of L2 regularization in Adam. The update rule is:

\begin{align}
m_t &= \beta_1 m_{t-1} + (1-\beta_1) g_t, \\
v_t &= \beta_2 v_{t-1} + (1-\beta_2) g_t^2, \\
\hat{m}_t &= \frac{m_t}{1-\beta_1^t}, \quad \hat{v}_t = \frac{v_t}{1-\beta_2^t}, \\
\theta_{t+1} &= \theta_t - \eta \left(\frac{\hat{m}_t}{\sqrt{\hat{v}_t} + \epsilon} + \lambda \theta_t\right),
\end{align}
where $\lambda$ is the weight decay coefficient.

\textbf{SGD with Momentum}
, Stochastic Gradient Descent with momentum accelerates convergence by accumulating a velocity vector in directions of persistent gradient reduction:

\begin{align}
v_t &= \mu v_{t-1} + g_t, \\
\theta_{t+1} &= \theta_t - \eta v_t,
\end{align}
where $\mu$ is the momentum coefficient (typically 0.9)

\textbf{RMSProp Optimizer},
Root Mean Square Propagation adapts the learning rate for each parameter by dividing the gradient by a running average of recent gradient magnitudes:
\begin{align}
v_t &= \beta v_{t-1} + (1-\beta) g_t^2, \\
\theta_{t+1} &= \theta_t - \frac{\eta}{\sqrt{v_t} + \epsilon} g_t,
\end{align}
where $\beta=0.9$ is the decay rate for the moving average.

\subsubsection{Model Complexity and Training Time}

The total number of trainable parameters of the proposed model is approximately  $5.2$ million, which is significantly lower compared to traditional CNN models such as ResNet50 ($23.5$M parameters). The Tesla T4 GPU had an average training time per epoch of $10$ seconds, and the model converged in 40 or 50 epochs. This shows not only computational efficiency, but also its applicability to real-time applications on resource-constrained and embedded systems.

\section{Result Analysis}
\label{section:results}

This section presents a complete evaluation of our Bangladeshi banknote recognition system. We tested it using five different datasets and various hyperparameter settings. The results demonstrate the effectiveness of the proposed hybrid model.
\subsection{Evaluation of Model Performance Across Optimizers and Learning Rates}

Table \ref{tab:allOptimizerlearningrate} provides a detailed comparison of four optimization algorithms—Adam, AdamW, RMSprop, and SGD—assessed across five distinct datasets: Primary Dataset-1, Primary Dataset-2, Custom Dataset-1, Custom Dataset-2, and Custom Dataset-3. The evaluation metrics encompass Accuracy (A), Precision (P), Recall (R), and F1-score (F1), all of which were measured during the Training, Validation, and Testing phases. Each optimizer was evaluated using four learning rates (0.0001, 0.001, 0.01, and 0.1) to assess their impact on model performance, thereby informing the selection of training strategies.

Furthermore, Table \ref{tab:allOptimizerlearningrate} demonstrates the capacity of optimizers like Adam and AdamW to effectively train models, irrespective of the underlying complexity.
The findings demonstrate that both Adam and AdamW consistently yield superior accuracy and F1-scores across all datasets and learning rates, thereby suggesting robust generalization capabilities and stable operational characteristics. In summation, the Adam optimizer, particularly with a learning rate of 0.0001, exhibited the most consistent performance and effective generalization across all datasets when compared to the other optimizers and learning rates. The adaptive learning-rate strategies inherent in Adam and AdamW facilitate their superior performance in managing diverse feature distributions and fluctuating background complexities, which contributes to their enhanced stability.
Conversely, while RMSprop demonstrates strong performance during the training phase, its validation and test accuracy exhibit a minor decline when applied to intricate datasets. In contrast, SGD's performance is notably influenced by the learning rate, resulting in considerable fluctuations, especially when dealing with visually demanding datasets; this implies that its efficacy is contingent upon both the chosen optimization algorithm and the inherent complexity of the dataset.

\begin{table*}[t]
\centering
\caption{Comparison of Model Performance Across Optimizers and Learning Rates}
\label{tab:allOptimizerlearningrate}
\adjustbox{width=\textwidth,center}{%
\begin{tabular}{lll|rrrr|rrrr|rrrr|rrrr|rrrr}
\hline
Optimizer & LR & Eval & \multicolumn{4}{c|}{Custom Dataset-1} & \multicolumn{4}{c|}{Primary Dataset-1} & \multicolumn{4}{c|}{Custom Dataset-2} & \multicolumn{4}{c|}{Primary Dataset-2} & \multicolumn{4}{c}{Custom Dataset-3} \\
\cline{4-23}
 &  &  & A & P & R & F1 & A & P & R & F1 & A & P & R & F1 & A & P & R & F1 & A & P & R & F1 \\
\hline
\multirow{3}{*}{Adam} & 0.0001 & Training & 1.0000 & 1.0000 & 1.0000 & 1.0000 & 1.0000 & 1.0000& 1.0000 & 1.0000 & 0.9999 & 0.9999 & 0.9999 & 0.9999  & 1.0000  & 1.0000  & 1.0000 & 1.0000 & 0.9999 & 0.9999 & 0.9999 & 0.9999  \\
& & Validation & 0.9967 & 0.9968 & 0.9967 & 0.9967 & 0.9869 & 0.9874 & 0.9869 & 0.9870 & 0.9899 & 0.9900 & 0.9899 & 0.9899 & 0.9717 & 0.9721 & 0.9717 & 0.9717 & 0.9680 & 0.9682 & 0.9680 & 0.9681 \\
& & Testing & 0.9975 & 0.9976 & 0.9975 & 0.9975 & 0.9795 & 0.9800 & 0.9795 & 0.9796 & 0.9866 & 0.9869 & 0.9866 & 0.9866 & 0.9284 & 0.9288 & 0.9284 & 0.9281 & 0.9498 & 0.9497 & 0.9498 & 0.9496 \\
& 0.001 & Training & 1.0000 & 1.0000 & 1.0000 & 1.0000 & 1.0000 & 1.0000 & 1.0000 & 1.0000 & 0.9999 & 0.9999 & 0.9999 & 0.9999 & 1.0000 & 1.0000 & 1.0000 & 1.0000 & 0.9999 & 0.9999 & 0.9999 & 0.9999 \\
& & Validation & 0.9935 & 0.9935 & 0.9935 & 0.9935 & 0.9895 & 0.9896 & 0.9895 & 0.9895 & 0.9911 & 0.9911 & 0.9911 & 0.9911 & 0.9575 & 0.9577 & 0.9575 & 0.9574 & 0.9746 & 0.9747 & 0.9746 & 0.9746 \\
& & Testing & 0.9975 & 0.9976 & 0.9975 & 0.9975 & 0.9795 & 0.9800 & 0.9795 & 0.9796 & 0.9866 & 0.9869 & 0.9866 & 0.9866 & 0.9284 & 0.9288 & 0.9284 & 0.9281 & 0.9432 & 0.9435 & 0.9432 & 0.9431 \\
& 0.01 & Training & 0.9997 & 0.9997 & 0.9997 & 0.9997 & 1.0000 & 1.0000 & 1.0000 & 1.0000 & 1.0000 & 1.0000 & 1.0000 & 1.0000 & 1.0000 & 1.0000 & 1.0000 & 1.0000 & 0.9999 & 0.9999 & 0.9999 & 0.9999 \\
& & Validation & 0.9967 & 0.9968 & 0.9967 & 0.9967 & 0.9843 & 0.9844 & 0.9843 & 0.9843 & 0.9935 & 0.9935 & 0.9935 & 0.9935 & 0.9706 & 0.9712 & 0.9706 & 0.9706 & 0.9695 & 0.9696 & 0.9695 & 0.9695 \\
& & Testing & 1.0000 & 1.0000 & 1.0000 & 1.0000 & 0.9737 & 0.9746 & 0.9737 & 0.9739 & 0.9839 & 0.9841 & 0.9839 & 0.9840 & 0.9185 & 0.9194 & 0.9185 & 0.9175 & 0.9366 & 0.9365 & 0.9366 & 0.9364 \\
& 0.1 & Training & 1.0000 & 1.0000 & 1.0000 & 1.0000 & 1.0000 & 1.0000 & 1.0000 & 1.0000 & 0.9999 & 0.9999 & 0.9999 & 0.9999 & 1.0000 & 1.0000 & 1.0000 & 1.0000 & 0.9999 & 0.9999 & 0.9999 & 0.9999 \\
& & Validation & 0.9956 & 0.9957 & 0.9956 & 0.9956 & 0.9882 & 0.9884 & 0.9882 & 0.9882 & 0.9893 & 0.9893 & 0.9893 & 0.9893 & 0.9575 & 0.9590 & 0.9575 & 0.9578 & 0.9622 & 0.9623 & 0.9622 & 0.9622 \\
& & Testing & 0.9926 & 0.9929 & 0.9926 & 0.9926 & 0.9737 & 0.9740 & 0.9737 & 0.9738 & 0.9853 & 0.9854 & 0.9853 & 0.9853 & 0.9210 & 0.9235 & 0.9210 & 0.9208 & 0.9333 & 0.9331 & 0.9333 & 0.9329 \\
\hline
\multirow{12}{*}{AdamW} & 0.0001 & Training   & 1.0000 & 1.0000 & 1.0000 & 1.0000 & 1.0000 & 1.0000 & 1.0000 & 1.0000 & 1.0000 & 1.0000 & 1.0000 & 1.0000 & 1.0000 & 1.0000 & 1.0000 & 1.0000 & 0.9998 & 0.9998 & 0.9998 & 0.9998 \\
      &        & Validation & 0.9967 & 0.9967 & 0.9967 & 0.9967 & 0.9856 & 0.9859 & 0.9856 & 0.9857 & 0.9905 & 0.9906 & 0.9905 & 0.9905 & 0.9662 & 0.9664 & 0.9662 & 0.9662 & 0.9648 & 0.9649 & 0.9648 & 0.9647 \\
      &        & Testing    & 1.0000 & 1.0000 & 1.0000 & 1.0000 & 0.9766 & 0.9769 & 0.9766 & 0.9767 & 0.9853 & 0.9854 & 0.9853 & 0.9853 & 0.9309 & 0.9326 & 0.9309 & 0.9297 & 0.9366 & 0.9368 & 0.9366 & 0.9364 \\  
      & 0.001  & Training   & 1.0000 & 1.0000 & 1.0000 & 1.0000 & 1.0000 & 1.0000 & 1.0000 & 1.0000 & 0.9997 & 0.9997 & 0.9997 & 0.9997 & 1.0000 & 1.0000 & 1.0000 & 1.0000 & 0.9999 & 0.9999 & 0.9999 & 0.9999 \\
      &        & Validation & 0.9967 & 0.9968 & 0.9967 & 0.9967 & 0.9908 & 0.9910 & 0.9908 & 0.9909 & 0.9923 & 0.9923 & 0.9923 & 0.9923 & 0.9739 & 0.9741 & 0.9739 & 0.9739 & 0.9746 & 0.9747 & 0.9746 & 0.9746 \\
      &        & Testing    & 1.0000 & 1.0000 & 1.0000 & 1.0000 & 0.9766 & 0.9771 & 0.9766 & 0.9766 & 0.9853 & 0.9855 & 0.9853 & 0.9853 & 0.9210 & 0.9226 & 0.9210 & 0.9199 & 0.9481 & 0.9483 & 0.9481 & 0.9481 \\  
      & 0.01   & Training   & 1.0000 & 1.0000 & 1.0000 & 1.0000 & 1.0000 & 1.0000 & 1.0000 & 1.0000 & 1.0000 & 1.0000 & 1.0000 & 1.0000 & 1.0000 & 1.0000 & 1.0000 & 1.0000 & 0.9994 & 0.9994 & 0.9994 & 0.9994 \\
      &        & Validation & 0.9946 & 0.9946 & 0.9946 & 0.9946 & 0.9869 & 0.9871 & 0.9869 & 0.9869 & 0.9935 & 0.9935 & 0.9935 & 0.9934 & 0.9673 & 0.9676 & 0.9673 & 0.9674 & 0.9590 & 0.9591 & 0.9590 & 0.9590 \\
      &        & Testing    & 0.9951 & 0.9952 & 0.9951 & 0.9950 & 0.9708 & 0.9713 & 0.9708 & 0.9708 & 0.9893 & 0.9894 & 0.9893 & 0.9893 & 0.9333 & 0.9336 & 0.9333 & 0.9326 & 0.9267 & 0.9275 & 0.9267 & 0.9267 \\  
      & 0.1    & Training   & 1.0000 & 1.0000 & 1.0000 & 1.0000 & 1.0000 & 1.0000 & 1.0000 & 1.0000 & 0.9988 & 0.9988 & 0.9988 & 0.9988 & 1.0000 & 1.0000 & 1.0000 & 1.0000 & 0.9999 & 0.9999 & 0.9999 & 0.9999 \\
      &        & Validation & 0.9946 & 0.9946 & 0.9946 & 0.9945 & 0.9869 & 0.9872 & 0.9869 & 0.9869 & 0.9851 & 0.9851 & 0.9851 & 0.9851 & 0.9673 & 0.9679 & 0.9673 & 0.9675 & 0.9633 & 0.9635 & 0.9633 & 0.9633 \\
      &        & Testing    & 0.9975 & 0.9976 & 0.9975 & 0.9975 & 0.9766 & 0.9771 & 0.9766 & 0.9767 & 0.9866 & 0.9867 & 0.9866 & 0.9866 & 0.9136 & 0.9140 & 0.9136 & 0.9132 & 0.9366 & 0.9372 & 0.9366 & 0.9366 \\
  \hline
\multirow{12}{*}{RMSprop} & 0.0001 & Training   & 1.0000 & 1.0000 & 1.0000 & 1.0000 & 1.0000 & 1.0000 & 1.0000 & 1.0000 & 0.9999 & 0.9999 & 0.9999 & 0.9999 & 1.0000 & 1.0000 & 1.0000 & 1.0000 & 0.9999 & 0.9999 & 0.9999 & 0.9999 \\
        &        & Validation & 0.9967 & 0.9968 & 0.9967 & 0.9967 & 0.9882 & 0.9884 & 0.9882 & 0.9882 & 0.9929 & 0.9929 & 0.9929 & 0.9929 & 0.9684 & 0.9691 & 0.9684 & 0.9686 & 0.9680 & 0.9681 & 0.9680 & 0.9680 \\
        &        & Testing    & 0.9951 & 0.9952 & 0.9951 & 0.9951 & 0.9766 & 0.9768 & 0.9766 & 0.9766 & 0.9879 & 0.9881 & 0.9879 & 0.9879 & 0.9259 & 0.9281 & 0.9259 & 0.9252 & 0.9366 & 0.9370 & 0.9366 & 0.9365 \\  
        & 0.001  & Training   & 1.0000 & 1.0000 & 1.0000 & 1.0000 & 1.0000 & 1.0000 & 1.0000 & 1.0000 & 0.9999 & 0.9999 & 0.9999 & 0.9999 & 1.0000 & 1.0000 & 1.0000 & 1.0000 & 0.9999 & 0.9999 & 0.9999 & 0.9999 \\
        &        & Validation & 0.9967 & 0.9968 & 0.9967 & 0.9967 & 0.9882 & 0.9886 & 0.9882 & 0.9883 & 0.9946 & 0.9947 & 0.9946 & 0.9947 & 0.9706 & 0.9713 & 0.9706 & 0.9707 & 0.9764 & 0.9765 & 0.9764 & 0.9764 \\
        &        & Testing    & 0.9975 & 0.9976 & 0.9975 & 0.9975 & 0.9766 & 0.9771 & 0.9766 & 0.9766 & 0.9879 & 0.9880 & 0.9879 & 0.9879 & 0.9284 & 0.9283 & 0.9284 & 0.9277 & 0.9457 & 0.9461 & 0.9457 & 0.9455 \\  
        & 0.01   & Training   & 1.0000 & 1.0000 & 1.0000 & 1.0000 & 1.0000 & 1.0000 & 1.0000 & 1.0000 & 1.0000 & 1.0000 & 1.0000 & 1.0000 & 1.0000 & 1.0000 & 1.0000 & 1.0000 & 0.9995 & 0.9995 & 0.9995 & 0.9995 \\
        &        & Validation & 0.9967 & 0.9968 & 0.9967 & 0.9967 & 0.9895 & 0.9897 & 0.9895 & 0.9895 & 0.9935 & 0.9935 & 0.9935 & 0.9934 & 0.9739 & 0.9743 & 0.9739 & 0.9739 & 0.9601 & 0.9604 & 0.9601 & 0.9601 \\
        &        & Testing    & 1.0000 & 1.0000 & 1.0000 & 1.0000 & 0.9854 & 0.9858 & 0.9854 & 0.9855 & 0.9893 & 0.9894 & 0.9893 & 0.9893 & 0.9185 & 0.9183 & 0.9185 & 0.9177 & 0.9267 & 0.9274 & 0.9267 & 0.9267 \\  
        & 0.1    & Training   & 1.0000 & 1.0000 & 1.0000 & 1.0000 & 0.9997 & 0.9997 & 0.9997 & 0.9997 & 0.9999 & 0.9999 & 0.9999 & 0.9999 & 1.0000 & 1.0000 & 1.0000 & 1.0000 & 0.9997 & 0.9997 & 0.9997 & 0.9997 \\
        &        & Validation & 0.9967 & 0.9967 & 0.9967 & 0.9967 & 0.9830 & 0.9833 & 0.9830 & 0.9829 & 0.9875 & 0.9875 & 0.9875 & 0.9875 & 0.9695 & 0.9698 & 0.9695 & 0.9695 & 0.9633 & 0.9635 & 0.9633 & 0.9634 \\
        &        & Testing    & 1.0000 & 1.0000 & 1.0000 & 1.0000 & 0.9708 & 0.9711 & 0.9708 & 0.9708 & 0.9933 & 0.9934 & 0.9933 & 0.9933 & 0.9358 & 0.9360 & 0.9358 & 0.9355 & 0.9366 & 0.9369 & 0.9366 & 0.9364 \\
  \hline
\multirow{12}{*}{SGD} & 0.0001 & Training   & 0.9894 & 0.9895 & 0.9894 & 0.9894 & 0.9522 & 0.9528 & 0.9522 & 0.9521 & 0.9756 & 0.9756 & 0.9756 & 0.9756 & 0.8881 & 0.8885 & 0.8881 & 0.8877 & 0.9177 & 0.9184 & 0.9177 & 0.9177 \\
    &        & Validation & 0.9760 & 0.9773 & 0.9760 & 0.9762 & 0.9188 & 0.9196 & 0.9188 & 0.9186 & 0.9524 & 0.9531 & 0.9524 & 0.9526 & 0.8192 & 0.8244 & 0.8192 & 0.8189 & 0.8751 & 0.8757 & 0.8751 & 0.8748 \\
    &        & Testing    & 0.9778 & 0.9783 & 0.9778 & 0.9779 & 0.9181 & 0.9244 & 0.9181 & 0.9180 & 0.9639 & 0.9647 & 0.9639 & 0.9638 & 0.8148 & 0.8194 & 0.8148 & 0.8156 & 0.8733 & 0.8729 & 0.8733 & 0.8727 \\  
    & 0.001  & Training   & 1.0000 & 1.0000 & 1.0000 & 1.0000 & 0.9993 & 0.9993 & 0.9993 & 0.9993 & 0.9991 & 0.9991 & 0.9991 & 0.9991 & 0.9992 & 0.9992 & 0.9992 & 0.9992 & 0.9975 & 0.9975 & 0.9975 & 0.9975 \\
    &        & Validation & 0.9967 & 0.9967 & 0.9967 & 0.9967 & 0.9804 & 0.9807 & 0.9804 & 0.9804 & 0.9839 & 0.9842 & 0.9839 & 0.9840 & 0.9564 & 0.9571 & 0.9564 & 0.9566 & 0.9503 & 0.9504 & 0.9503 & 0.9502 \\
    &        & Testing    & 0.9975 & 0.9976 & 0.9975 & 0.9975 & 0.9795 & 0.9801 & 0.9795 & 0.9796 & 0.9879 & 0.9880 & 0.9879 & 0.9879 & 0.9235 & 0.9233 & 0.9235 & 0.9226 & 0.9317 & 0.9322 & 0.9317 & 0.9315 \\  
    & 0.01   & Training   & 1.0000 & 1.0000 & 1.0000 & 1.0000 & 1.0000 & 1.0000 & 1.0000 & 1.0000 & 0.9991 & 0.9991 & 0.9991 & 0.9991 & 0.9997 & 0.9997 & 0.9997 & 0.9997 & 0.9926 & 0.9926 & 0.9926 & 0.9926 \\
    &        & Validation & 0.9967 & 0.9967 & 0.9967 & 0.9967 & 0.9804 & 0.9810 & 0.9804 & 0.9804 & 0.9834 & 0.9835 & 0.9834 & 0.9834 & 0.9586 & 0.9592 & 0.9586 & 0.9588 & 0.9394 & 0.9393 & 0.9394 & 0.9393 \\
    &        & Testing    & 0.9951 & 0.9953 & 0.9951 & 0.9951 & 0.9825 & 0.9827 & 0.9825 & 0.9825 & 0.9866 & 0.9868 & 0.9866 & 0.9866 & 0.9235 & 0.9237 & 0.9235 & 0.9228 & 0.9193 & 0.9190 & 0.9193 & 0.9187 \\  
    & 0.1    & Training   & 0.9997 & 0.9997 & 0.9997 & 0.9997 & 0.9997 & 0.9997 & 0.9997 & 0.9997 & 0.9990 & 0.9990 & 0.9990 & 0.9990 & 0.9995 & 0.9995 & 0.9995 & 0.9995 & 0.9900 & 0.9900 & 0.9900 & 0.9900 \\
    &        & Validation & 0.9946 & 0.9946 & 0.9946 & 0.9945 & 0.9817 & 0.9821 & 0.9817 & 0.9817 & 0.9869 & 0.9871 & 0.9869 & 0.9869 & 0.9597 & 0.9600 & 0.9597 & 0.9597 & 0.9310 & 0.9312 & 0.9310 & 0.9310 \\
    &        & Testing    & 0.9951 & 0.9952 & 0.9951 & 0.9951 & 0.9766 & 0.9773 & 0.9766 & 0.9768 & 0.9893 & 0.9894 & 0.9893 & 0.9893 & 0.9259 & 0.9270 & 0.9259 & 0.9251 & 0.9119 & 0.9119 & 0.9119 & 0.9114 \\
\hline
\end{tabular}%
}
\vspace{0.1cm}
\small
\\ \textbf{Abbreviations:} LR= Learning Rate, Eval = Evaluation, A=Accuracy, P=Precision, R=Recall, F1=F1-score
\vspace{0.1cm}
\end{table*}

It is important to note that the proposed model has near-perfect training accuracy in most optimizer and learning rate configurations. This shows the hybrid architecture's representational ability, not memorization. Indeed, validation and test performance are consistently worse and vary across datasets of various complexity, demonstrating that the model is not just overfitted to the training data.

\subsection{Comparative Performance Across Datasets}

The subsequent discussion details the experimental results derived from the five selected datasets, encompassing the cross-validation outcomes and the results of a thorough evaluation using metrics such as Accuracy, Precision, Recall, F1-score, Cohen's Kappa coefficient, Matthews Correlation Coefficient (MCC), and Area Under the ROC Curve (AUC). To ensure the statistical robustness of the findings and to mitigate potential bias, each experimental run underwent cross-validation.

\subsubsection{Cross-Validation Performance Analysis}

Table~\ref{tab:cross_validation_simplified} summarizes the performance of five-fold cross-validation across all datasets. The model shows strong generalization across the datasets, as indicated by high evaluation and agreement metrics, which suggests consistent discrimination.

The model performance degrades in relatively complex and challenging visual environments like \textbf{Primary Dataset-2}; however, the agreement metrics remain above 0.92 ($\kappa$ = 0.9211, MCC = 0.9213). It clearly demonstrates that cluttered, uncontrolled backgrounds pose a difficult challenge. However, the model also has high AUC performance (0.9906), which shows that it can obtain discriminative ability in spite of the complexity of visual information. Taken together, these findings are taken to show that the model is highly reliable and highly discriminative for all levels of visual complexity.

\begin{table*}[htbp]
\caption{Cross-Validation Results: Validation Accuracy and Test Metrics (\%)}
\label{tab:cross_validation_simplified}
\renewcommand{\arraystretch}{0.8}
\adjustbox{width=\textwidth,center,keepaspectratio}{%
\begin{tabular}{l|c|c|ccccccc}
\hline
\multirow{2}{*}{\textbf{Dataset}} & \multirow{2}{*}{\textbf{Fold}}
& \textbf{Val.}
& \multicolumn{7}{c}{\textbf{Test Metrics}} \\ \cline{4-10}
 &  & Acc. & Acc. & Prec. & Rec. & F1 & $\kappa$ & MCC & AUC \\
\hline
\multirow{8}{*}{\textbf{Custom Dataset-1}}
& 1 & 99.89 & 99.01 & 99.06 & 99.01 & 99.01 & 98.89 & 98.90 & 100.00 \\
& 2 & 100.00 & 99.51 & 99.53 & 99.51 & 99.51 & 99.44 & 99.45 & 100.00 \\
& 3 & \cellcolor{lightblue}\textcolor{skyblue}{\textbf{99.89}} & \cellcolor{lightblue}\textcolor{skyblue}{\textbf{99.75}} & \cellcolor{lightblue}\textcolor{skyblue}{\textbf{99.76}} & \cellcolor{lightblue}\textcolor{skyblue}{\textbf{99.75}} & \cellcolor{lightblue}\textcolor{skyblue}{\textbf{99.75}} & \cellcolor{lightblue}\textcolor{skyblue}{\textbf{99.72}} & \cellcolor{lightblue}\textcolor{skyblue}{\textbf{99.72}} & \cellcolor{lightblue}\textcolor{skyblue}{\textbf{100.00}} \\
& 4 & 99.78 & 99.51 & 99.53 & 99.51 & 99.51 & 99.44 & 99.45 & 100.00 \\
& 5 & 99.46 & 99.51 & 99.51 & 99.51 & 99.51 & 99.44 & 99.44 & 100.00 \\
\cline{2-10}
& \textbf{Mean} & \textbf{99.80} & \textbf{99.46} & \textbf{99.48} & \textbf{99.46} & \textbf{99.46} & \textbf{99.39} & \textbf{99.39} & \textbf{100.00} \\
& \textbf{Std} & \textbf{0.21} & \textbf{0.27} & \textbf{0.25} & \textbf{0.27} & \textbf{0.27} & \textbf{0.30} & \textbf{0.30} & \textbf{0.00} \\
\hline

\multirow{8}{*}{\textbf{Primary Dataset-1}}
& 1 & 98.04 & 97.95 & 98.01 & 97.95 & 97.94 & 97.70 & 97.71 & 99.98 \\
& 2 & 98.69 & 98.25 & 98.26 & 98.25 & 98.25 & 98.03 & 98.03 & 99.98 \\
& 3 & 98.95 & 97.95 & 98.07 & 97.95 & 97.96 & 97.70 & 97.71 & 99.96 \\
& 4 & \cellcolor{lightblue}\textcolor{skyblue}{\textbf{98.43}} & \cellcolor{lightblue}\textcolor{skyblue}{\textbf{99.12}} & \cellcolor{lightblue}\textcolor{skyblue}{\textbf{99.15}} & \cellcolor{lightblue}\textcolor{skyblue}{\textbf{99.12}} & \cellcolor{lightblue}\textcolor{skyblue}{\textbf{99.12}} & \cellcolor{lightblue}\textcolor{skyblue}{\textbf{99.01}} & \cellcolor{lightblue}\textcolor{skyblue}{\textbf{99.02}} & \cellcolor{lightblue}\textcolor{skyblue}{\textbf{99.98}} \\
& 5 & 98.95 & 98.54 & 98.60 & 98.54 & 98.55 & 98.36 & 98.36 & 99.96 \\
\cline{2-10}
& \textbf{Mean} & \textbf{98.61} & \textbf{98.36} & \textbf{98.42} & \textbf{98.36} & \textbf{98.37} & \textbf{98.16} & \textbf{98.16} & \textbf{99.97} \\
& \textbf{Std} & \textbf{0.39} & \textbf{0.49} & \textbf{0.47} & \textbf{0.49} & \textbf{0.49} & \textbf{0.55} & \textbf{0.55} & \textbf{0.01} \\
\hline

\multirow{8}{*}{\textbf{Primary Dataset-2}}
& 1 & \cellcolor{lightblue}\textcolor{skyblue}{\textbf{97.39}} & \cellcolor{lightblue}\textcolor{skyblue}{\textbf{93.58}} & \cellcolor{lightblue}\textcolor{skyblue}{\textbf{93.70}} & \cellcolor{lightblue}\textcolor{skyblue}{\textbf{93.58}} & \cellcolor{lightblue}\textcolor{skyblue}{\textbf{93.55}} & \cellcolor{lightblue}\textcolor{skyblue}{\textbf{92.78}} & \cellcolor{lightblue}\textcolor{skyblue}{\textbf{92.80}} & \cellcolor{lightblue}\textcolor{skyblue}{\textbf{99.20}} \\
& 2 & 95.86 & 92.35 & 92.37 & 92.35 & 92.30 & 91.39 & 91.41 & 98.96 \\
& 3 & 97.39 & 93.83 & 93.90 & 93.83 & 93.78 & 93.06 & 93.08 & 99.08 \\
& 4 & 97.06 & 93.09 & 93.15 & 93.09 & 93.03 & 92.22 & 92.24 & 99.03 \\
& 5 & 96.08 & 92.10 & 92.14 & 92.10 & 92.07 & 91.11 & 91.12 & 99.05 \\
\cline{2-10}
& \textbf{Mean} & \textbf{96.75} & \textbf{92.99} & \textbf{93.05} & \textbf{92.99} & \textbf{92.95} & \textbf{92.11} & \textbf{92.13} & \textbf{99.06} \\
& \textbf{Std} & \textbf{0.73} & \textbf{0.75} & \textbf{0.78} & \textbf{0.75} & \textbf{0.75} & \textbf{0.85} & \textbf{0.85} & \textbf{0.09} \\
\hline

\multirow{8}{*}{\textbf{Custom Dataset-2}}
& 1 & 99.23 & 97.99 & 98.04 & 97.99 & 98.00 & 97.74 & 97.74 & 99.93 \\
& 2 & 98.87 & 98.66 & 98.68 & 98.66 & 98.66 & 98.49 & 98.50 & 99.98 \\
& 3 & 98.81 & 98.93 & 98.94 & 98.93 & 98.93 & 98.80 & 98.80 & 99.97 \\
& 4 & 99.46 & 98.39 & 98.41 & 98.39 & 98.39 & 98.19 & 98.19 & 99.97 \\
& 5 & \cellcolor{lightblue}\textcolor{skyblue}{\textbf{99.11}} & \cellcolor{lightblue}\textcolor{skyblue}{\textbf{99.06}} & \cellcolor{lightblue}\textcolor{skyblue}{\textbf{99.06}} & \cellcolor{lightblue}\textcolor{skyblue}{\textbf{99.06}} & \cellcolor{lightblue}\textcolor{skyblue}{\textbf{99.06}} & \cellcolor{lightblue}\textcolor{skyblue}{\textbf{98.95}} & \cellcolor{lightblue}\textcolor{skyblue}{\textbf{98.95}} & \cellcolor{lightblue}\textcolor{skyblue}{\textbf{99.97}} \\
\cline{2-10}
& \textbf{Mean} & \textbf{99.10} & \textbf{98.61} & \textbf{98.62} & \textbf{98.61} & \textbf{98.61} & \textbf{98.43} & \textbf{98.44} & \textbf{99.96} \\
& \textbf{Std} & \textbf{0.27} & \textbf{0.43} & \textbf{0.42} & \textbf{0.43} & \textbf{0.43} & \textbf{0.48} & \textbf{0.48} & \textbf{0.02} \\
\hline

\multirow{8}{*}{\textbf{Custom Dataset-3}}
& 1 & \cellcolor{lightblue}\textcolor{skyblue}{\textbf{97.17}} & \cellcolor{lightblue}\textcolor{skyblue}{\textbf{94.49}} & \cellcolor{lightblue}\textcolor{skyblue}{\textbf{94.53}} & \cellcolor{lightblue}\textcolor{skyblue}{\textbf{94.49}} & \cellcolor{lightblue}\textcolor{skyblue}{\textbf{94.48}} & \cellcolor{lightblue}\textcolor{skyblue}{\textbf{93.80}} & \cellcolor{lightblue}\textcolor{skyblue}{\textbf{93.80}} & \cellcolor{lightblue}\textcolor{skyblue}{\textbf{99.71}} \\
& 2 & 97.02 & 93.58 & 93.62 & 93.58 & 93.55 & 92.78 & 92.79 & 99.68 \\
& 3 & 96.70 & 93.42 & 93.44 & 93.42 & 93.39 & 92.59 & 92.60 & 99.68 \\
& 4 & 96.99 & 94.24 & 94.22 & 94.24 & 94.21 & 93.52 & 93.52 & 99.69 \\
& 5 & 96.08 & 93.58 & 93.63 & 93.58 & 93.58 & 92.78 & 92.78 & 99.70 \\
\cline{2-10}
& \textbf{Mean} & \textbf{96.79} & \textbf{93.86} & \textbf{93.89} & \textbf{93.86} & \textbf{93.84} & \textbf{93.09} & \textbf{93.10} & \textbf{99.69} \\
& \textbf{Std} & \textbf{0.43} & \textbf{0.47} & \textbf{0.47} & \textbf{0.47} & \textbf{0.47} & \textbf{0.53} & \textbf{0.53} & \textbf{0.02} \\
\hline

\end{tabular}
}
\vspace{0.1cm}
\small
\\ \textbf{Abbreviations:} Val=Validation, Acc=Accuracy, Prec=Precision, Rec=Recall, F1=F1-score, $\kappa$=Cohen Kappa Coefficient, MCC=Matthews Correlation Coefficient, AUC=Area Under Curve
\vspace{0.1cm}
\end{table*}

\subsubsection{Final Model Performance}
Table~\ref{tab:final_performance_all} summarizes the final model's performance across all datasets. The model shows consistent performance in both controlled and uncontrolled environments. It maintains stable agreement metrics even in more challenging real-world conditions. Although performance declines with visually complex datasets, the agreement measures and AUC stay stable across all datasets.

\begin{table*}[htbp]
\centering
\caption{Final Model Performance Across All Datasets (\%)}
\label{tab:final_performance_all}
\adjustbox{width=\textwidth,center,keepaspectratio}{%
\begin{tabular}{l|c|c|c|c|c}
\hline
\textbf{Metric} 
& \textbf{Custom Dataset-1} 
& \textbf{Primary Dataset-1} 
& \textbf{Primary Dataset-2}
& \textbf{Custom Dataset-2}
& \textbf{Custom Dataset-3} \\
\hline
Accuracy      & 99.75 & 97.95 & 92.84 & 98.66 & 94.98 \\
Precision     & 99.76 & 98.00 & 92.88 & 98.69 & 94.97 \\
Recall        & 99.75 & 97.95 & 92.84 & 98.66 & 94.98 \\
F1-Score      & 99.75 & 97.96 & 92.81 & 98.66 & 94.96 \\
Cohen's $\kappa$ & 99.72 & 97.70 & 91.94 & 98.49 & 94.35 \\
MCC           & 99.72 & 97.70 & 91.96 & 98.50 & 94.36 \\
AUC-ROC       & 100.00 & 99.94 & 99.15 & 99.98 & 99.74 \\
\hline
\end{tabular}}
\end{table*}

Performance on \textbf{Primary Dataset-3} exhibited a drop, because of more complex background elements, resulting in an accuracy of 92.84\% and an F1-score of 92.81\%.  Despite this, the model showed a strong agreement metric. It presented a $\kappa$ value of 0.9194 and an MCC of 0.9196 on the \textbf{Primary Dataset-2}. Furthermore, \textbf{Custom Dataset-2} yielded an accuracy of 0.9498 and an F1-score of 0.9496, coupled with solid $\kappa$ (0.9435) and MCC (0.9436) values.

The consistently elevated AUC-ROC values, ranging from 0.9915 to 1.0000, across all datasets, suggest a strong capacity for class separation. Consequently, these results corroborate the model's efficacy and stability across diverse imaging conditions, thereby preserving a high level of discriminative capability even within intricate environments.

\subsubsection{Confusion Matrix Analysis}

\begin{figure*}[htbp]
    \centering
    \begin{subfigure}[b]{0.48\textwidth}
        \includegraphics[width=\textwidth]{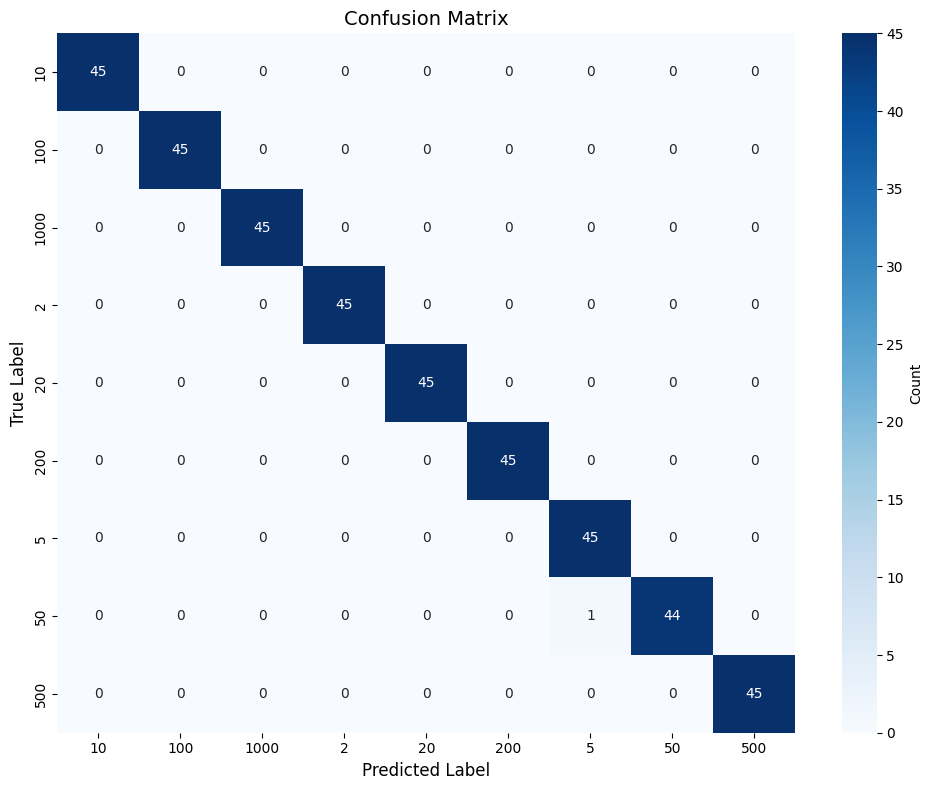}
        \caption{Primary Dataset-1}
    \end{subfigure}\hfill
    \begin{subfigure}[b]{0.48\textwidth}
        \includegraphics[width=\textwidth]{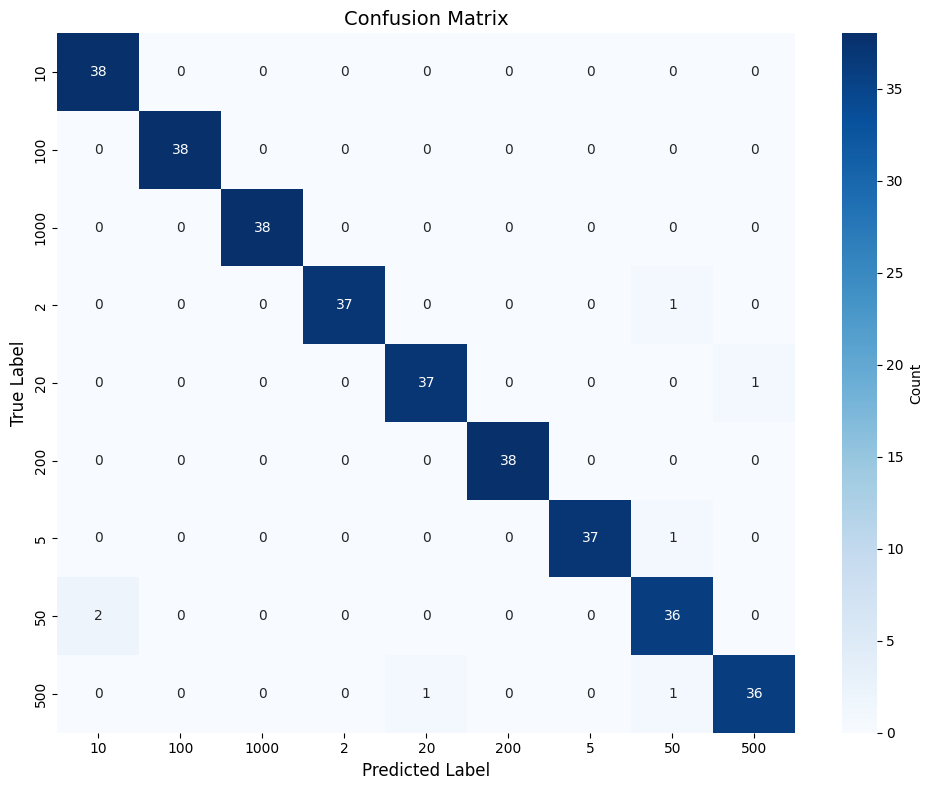}
        \caption{Primary Dataset-2}
    \end{subfigure}
    
    \vspace{0.4cm}
    
    \begin{subfigure}[b]{0.48\textwidth}
        \includegraphics[width=\textwidth]{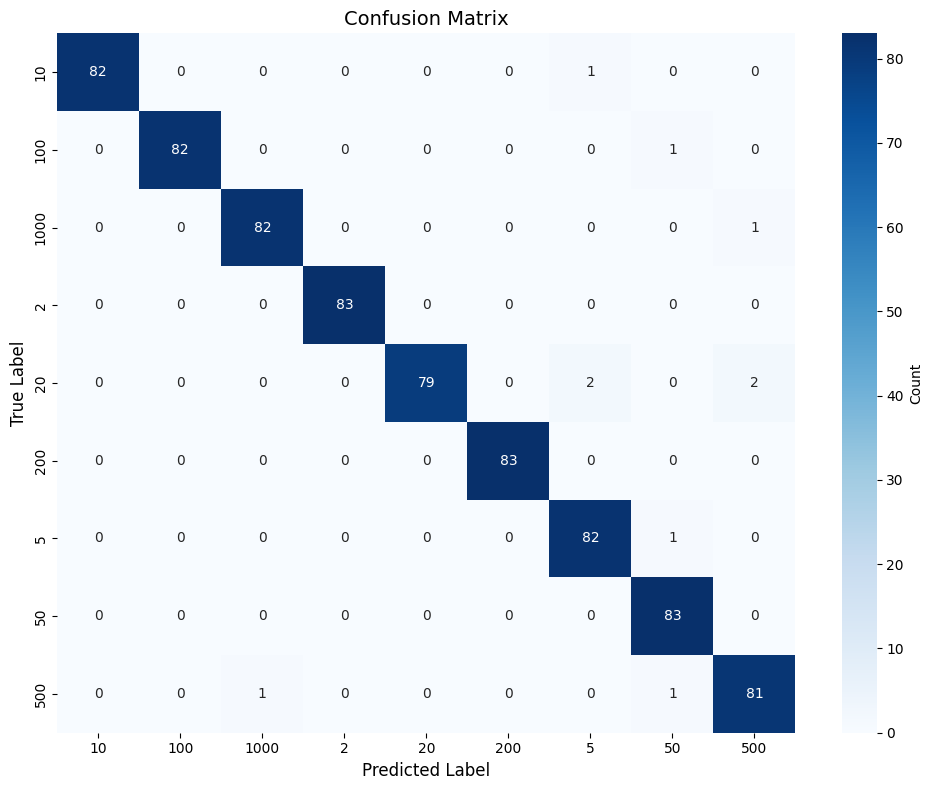}
        \caption{Custom Dataset-1}
    \end{subfigure}\hfill
    \begin{subfigure}[b]{0.48\textwidth}
        \includegraphics[width=\textwidth]{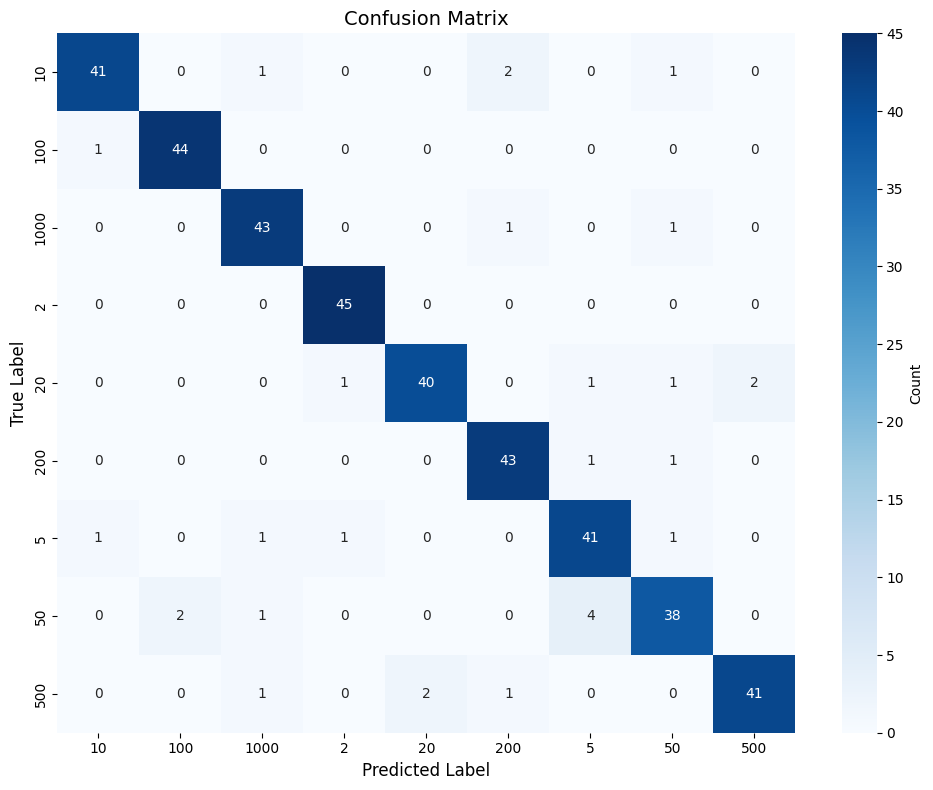}
        \caption{Primary Dataset-3}
    \end{subfigure}
    
    \vspace{0.4cm}
    
    \begin{subfigure}[b]{0.48\textwidth}
        \includegraphics[width=\textwidth]{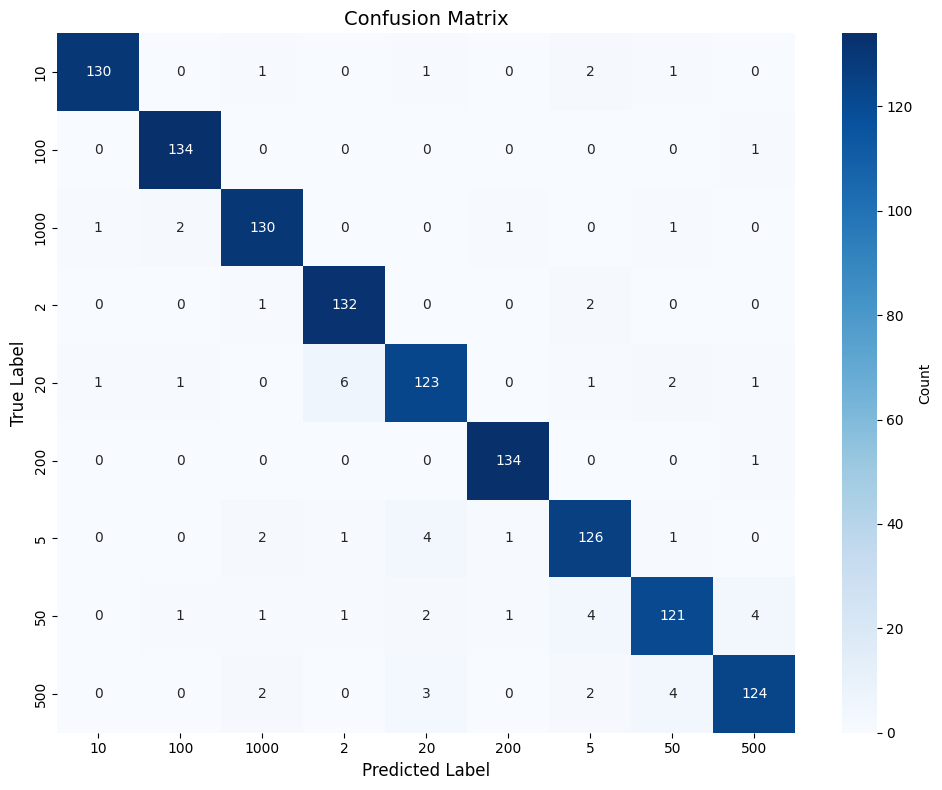}
        \caption{Custom Dataset-2}
    \end{subfigure}
    
    \caption{Test confusion matrices across all five datasets.}
    \label{fig:confusion_all}
\end{figure*}

Confusion matrices summarize classification behavior by comparing true labels (rows) to predicted labels (columns), with diagonal entries indicating correct predictions and off-diagonal entries indicating misclassifications. Figure~\ref{fig:confusion_all} shows the confusion matrices obtained across all datasets.

The overall accuracy reaches 99.75\% for \textbf{Primary Dataset-1}. On the other hand, per-class accuracies range from 95–100\% on \textbf{Primary Dataset-2} with scattered minor errors showing effective performance under moderate intra-class variability. Diagonal entries for each class are between 79-83, and macro-averaged accuracy is over 99\%, confirming good generalization on \textbf{Custom Dataset-1}. \textbf{Primary Dataset-3} fares just as well, with per-class accuracies of 93-99\%, while the sparse, non-systematic misclassifications suggest resistance to occlusion, clutter, illumination variation, and geometric distortion. Finally, the model highlights its capability of distinguishing and maintaining near state-of-the-art performance by exhibiting diagonal counts of 121–134 per class with least off-diagonal misclassifications on \textbf{Custom Dataset-2}. 

Altogether, the results show that the model displays near-perfect classification with minimal confusion and very robust across diverse datasets, further underscoring its readiness for practical applications in the real world.

\subsubsection{ROC-AUC Analysis}

An additional perspective on the model's ability to distinguish among classes across all datasets comes from the Receiver Operating Characteristic (ROC) and the corresponding Area Under the Curve (AUC). 

For \textbf{Primary Dataset-1} and \textbf{Primary Dataset-2}, the model achieves high per-class AUCs. This indicates solid feature learning despite moderate intra-class variability. In \textbf{Custom Dataset-1}, the model maintains consistently high per-class AUCs of 0.999 to 1.000. This shows excellent generalization across images with controlled or uniform backgrounds. In the more challenging \textbf{Primary Dataset-3}, AUCs range from 0.996 to 0.9998, with a micro-average of 0.9992. This indicates that the model can still achieve high true positive rates with very few false positives. Additionally, the model shows per-class AUCs of 0.995 to 1.000 in \textbf{Custom Dataset-2}, with micro- and macro-average AUCs of 0.998 and 0.997. This confirms reliable discrimination even among images of varying complexity.

\begin{figure*}[htbp]
    \centering
    \begin{subfigure}[b]{0.48\textwidth}
        \includegraphics[width=\textwidth]{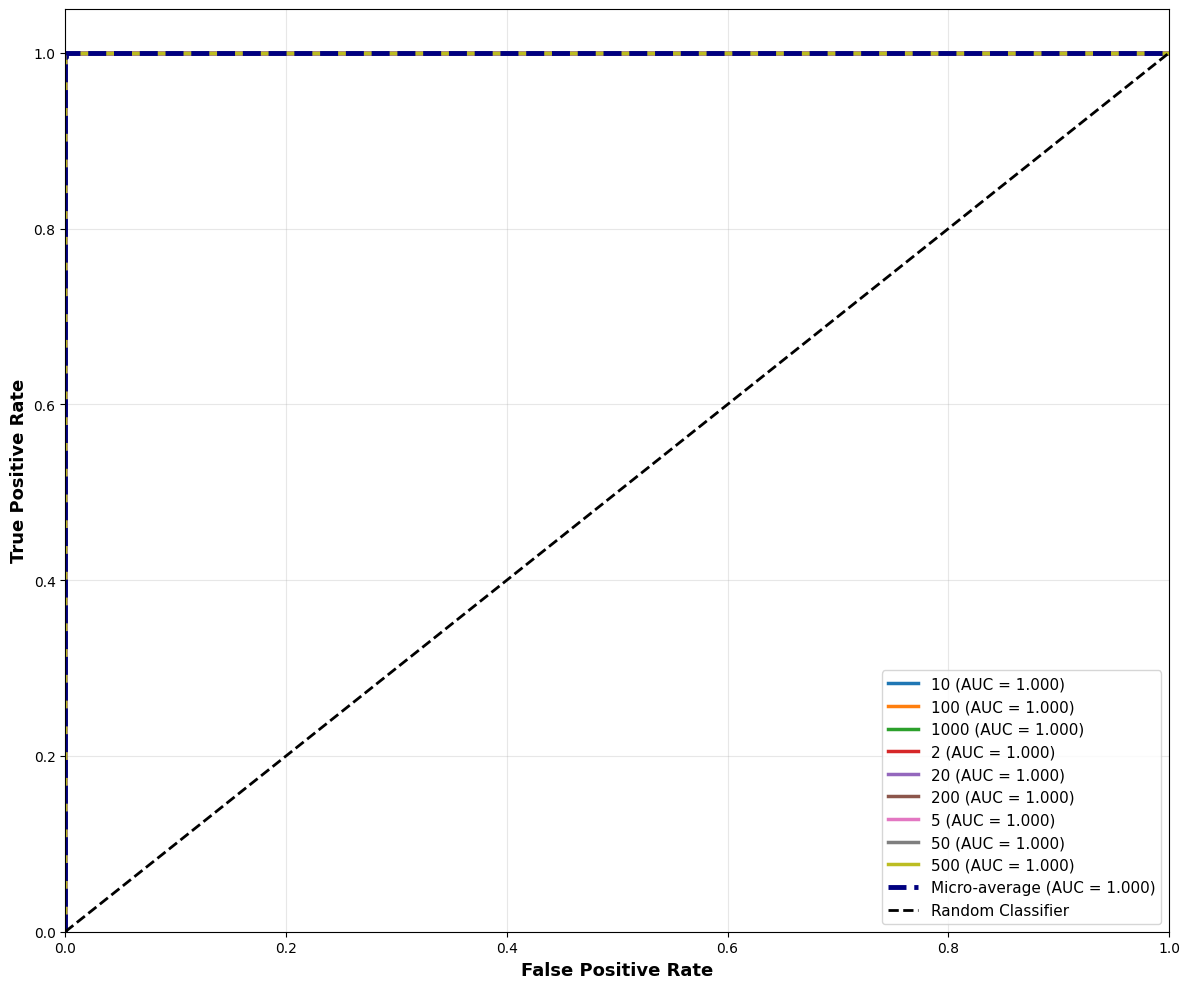}
        \caption{Primary Dataset-1}
    \end{subfigure}\hfill
    \begin{subfigure}[b]{0.48\textwidth}
        \includegraphics[width=\textwidth]{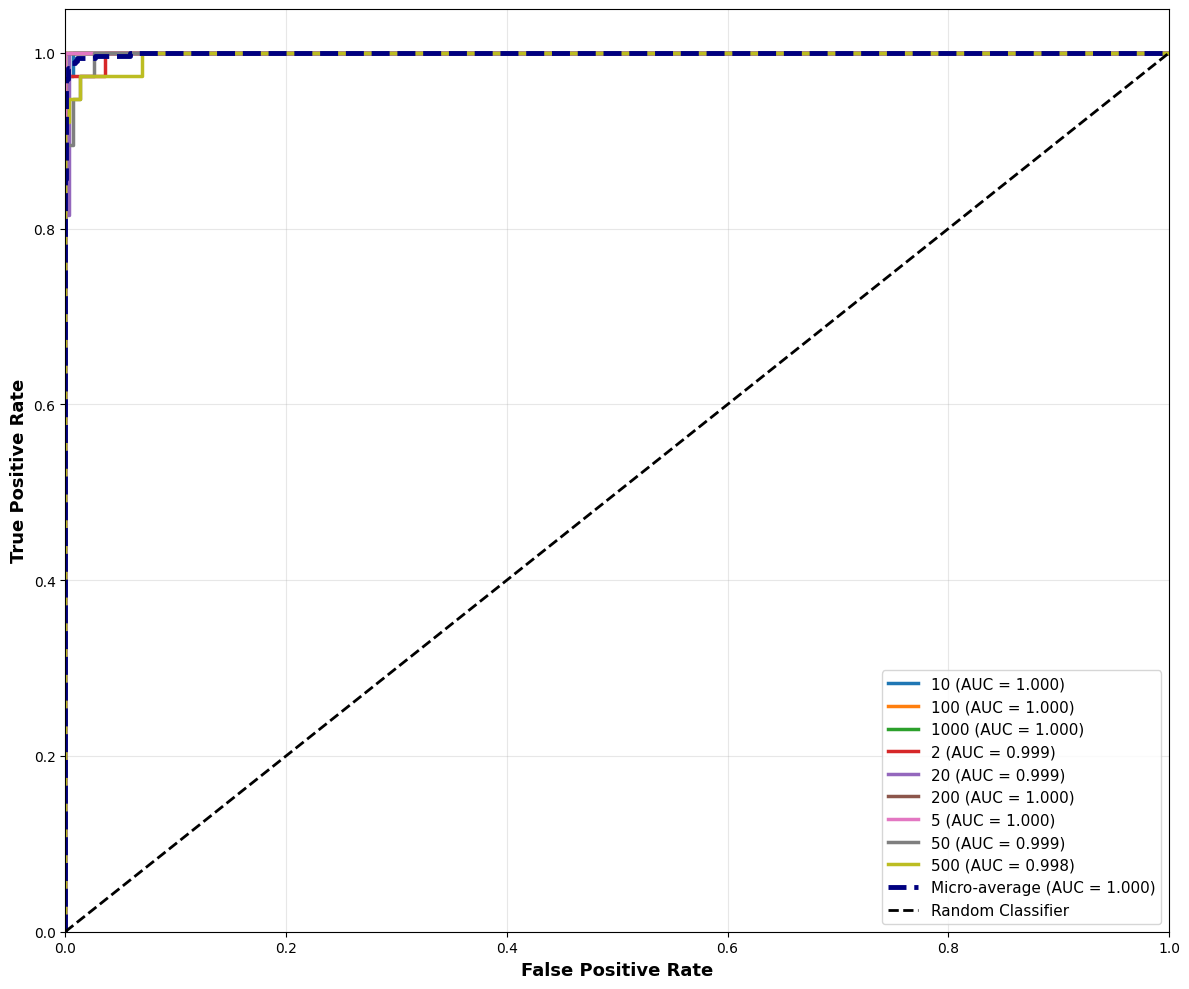}
        \caption{Primary Dataset-2}
    \end{subfigure}
    
    \vspace{0.4cm}
    
    \begin{subfigure}[b]{0.48\textwidth}
        \includegraphics[width=\textwidth]{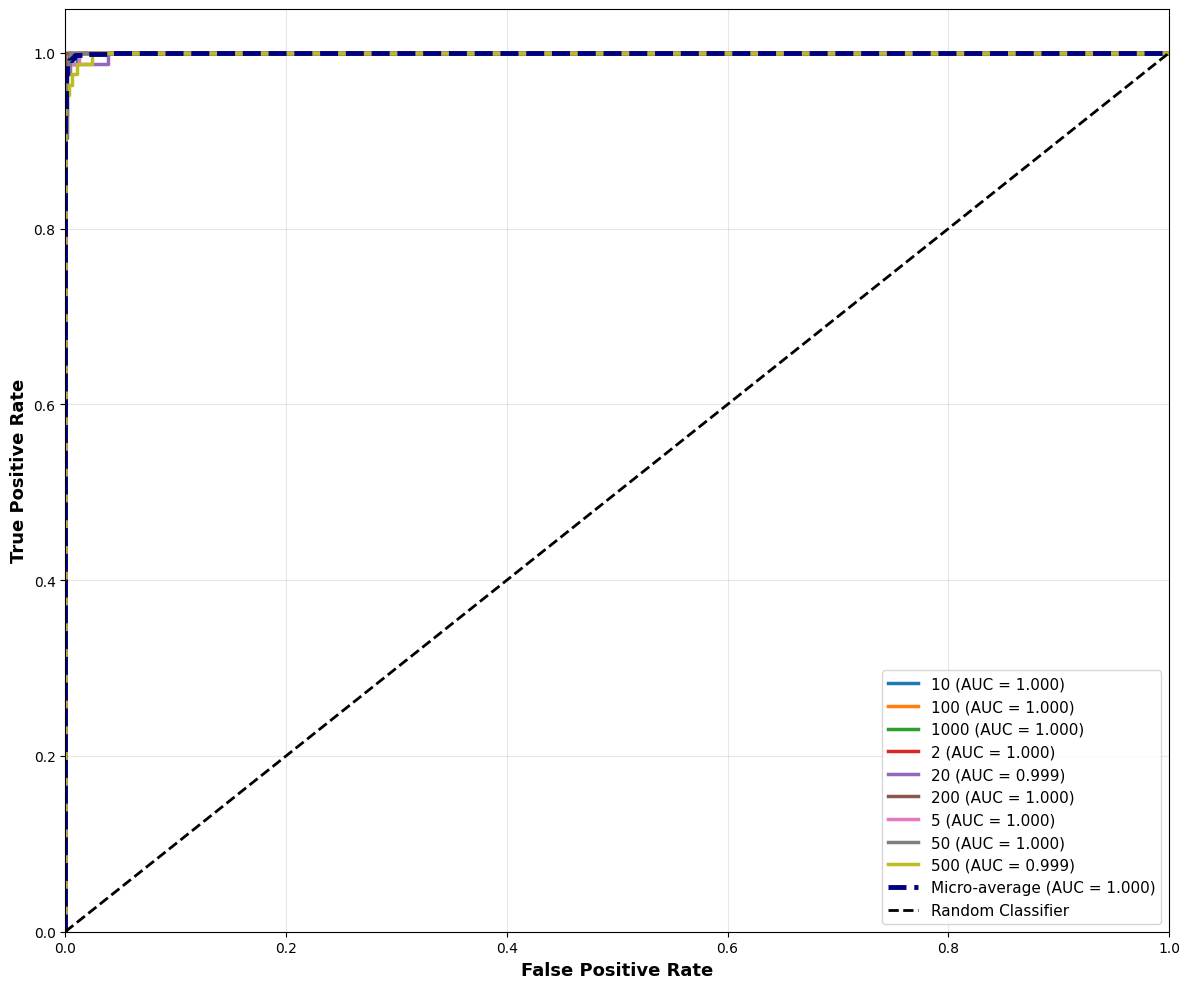}
        \caption{Custom Dataset-1}
    \end{subfigure}\hfill
    \begin{subfigure}[b]{0.48\textwidth}
        \includegraphics[width=\textwidth]{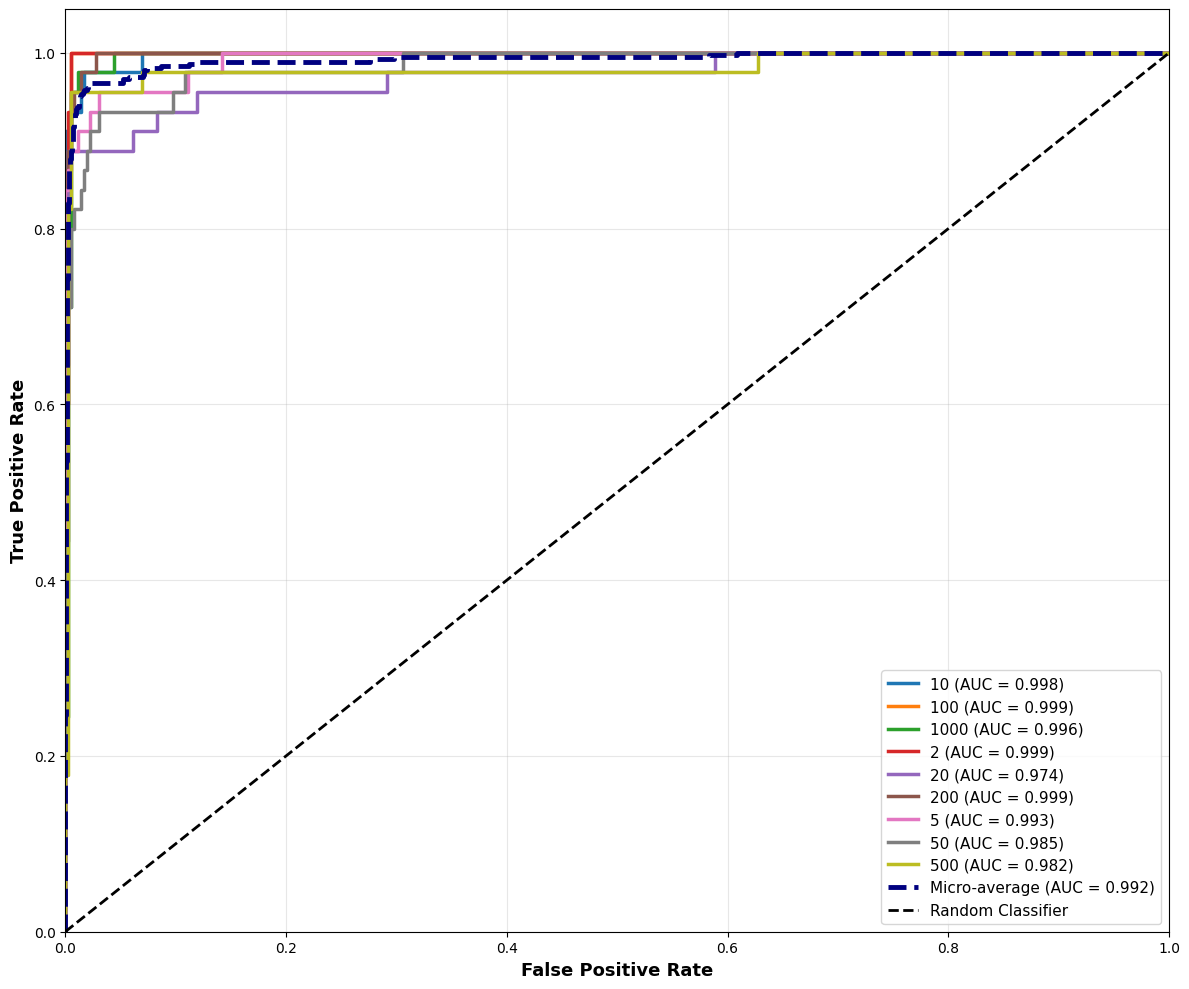}
        \caption{Primary Dataset-3}
    \end{subfigure}
    
    \vspace{0.4cm}
    
    \begin{subfigure}[b]{0.48\textwidth}
        \includegraphics[width=\textwidth]{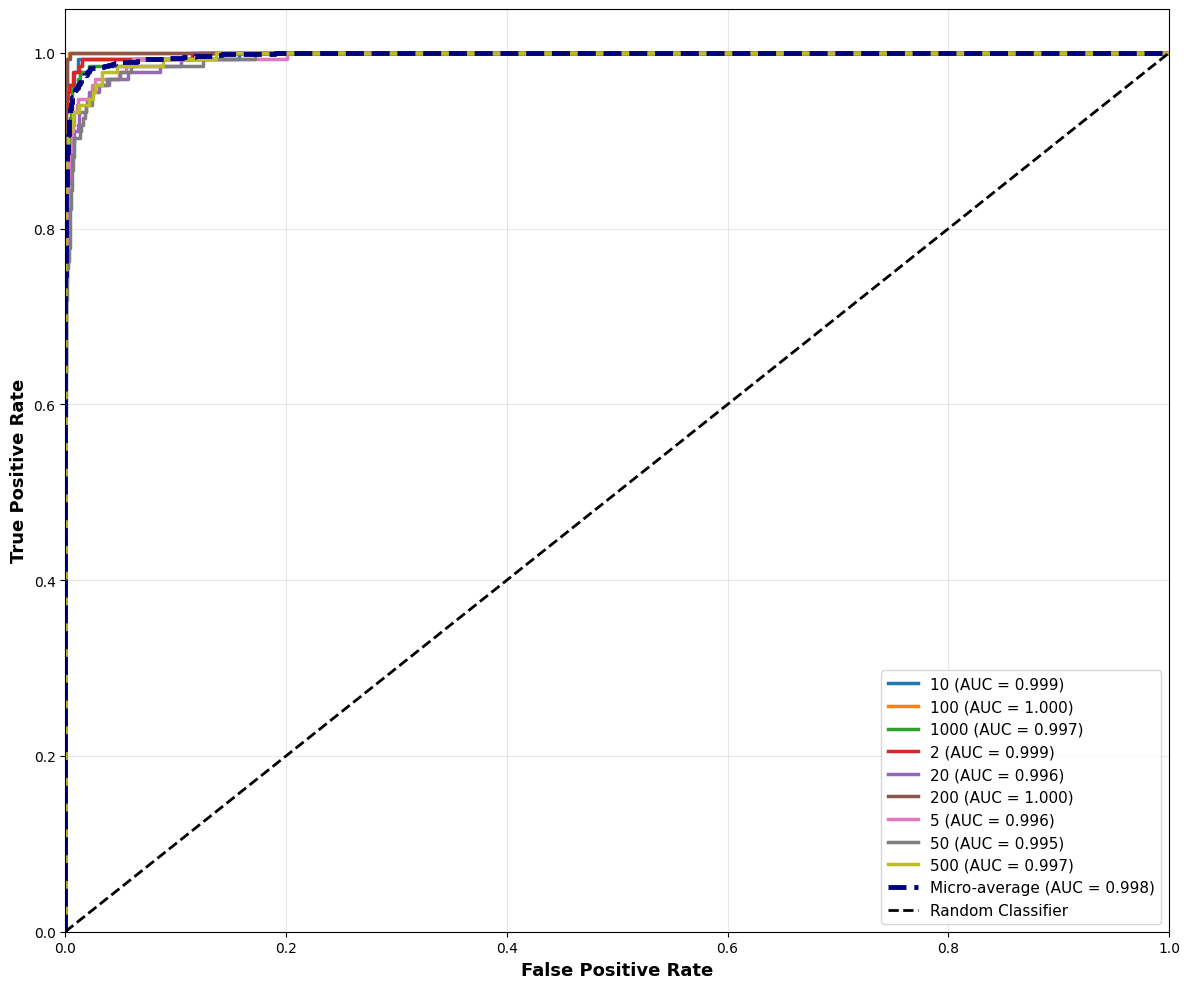}
        \caption{Custom Dataset-2}
    \end{subfigure}
    
    \caption{Multi-class ROC curves and AUC scores across five datasets.}
    \label{fig:roc_all_datasets}
\end{figure*}

Overall, the ROC-AUC, as shown in Figure~\ref {fig:roc_all_datasets}, results support the confusion matrix analysis. They highlight the model's well-calibrated confidence, strong class separation, and strong generalization in diverse real-world conditions.

\subsection{Model Interpretability Analysis}

To analyze model interpretability and provide a clear understanding of the model's decision-making process, XAI techniques such as LIME and SHAP were applied.

\subsubsection{Local Interpretability via LIME}

\begin{figure*}[htbp]
    \centering
    \includegraphics[width=0.32\linewidth]{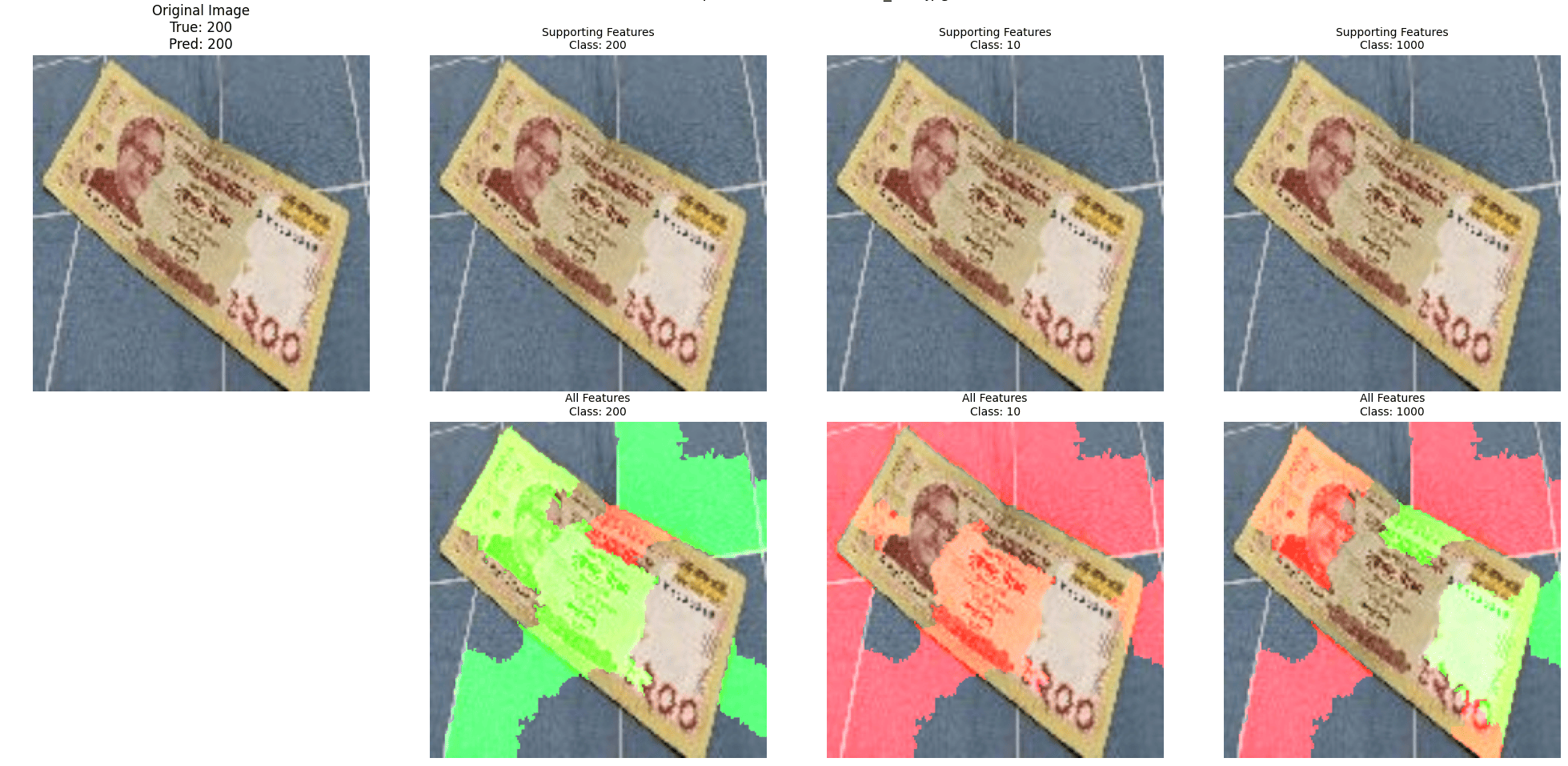}
    \hfill
    \includegraphics[width=0.32\linewidth]{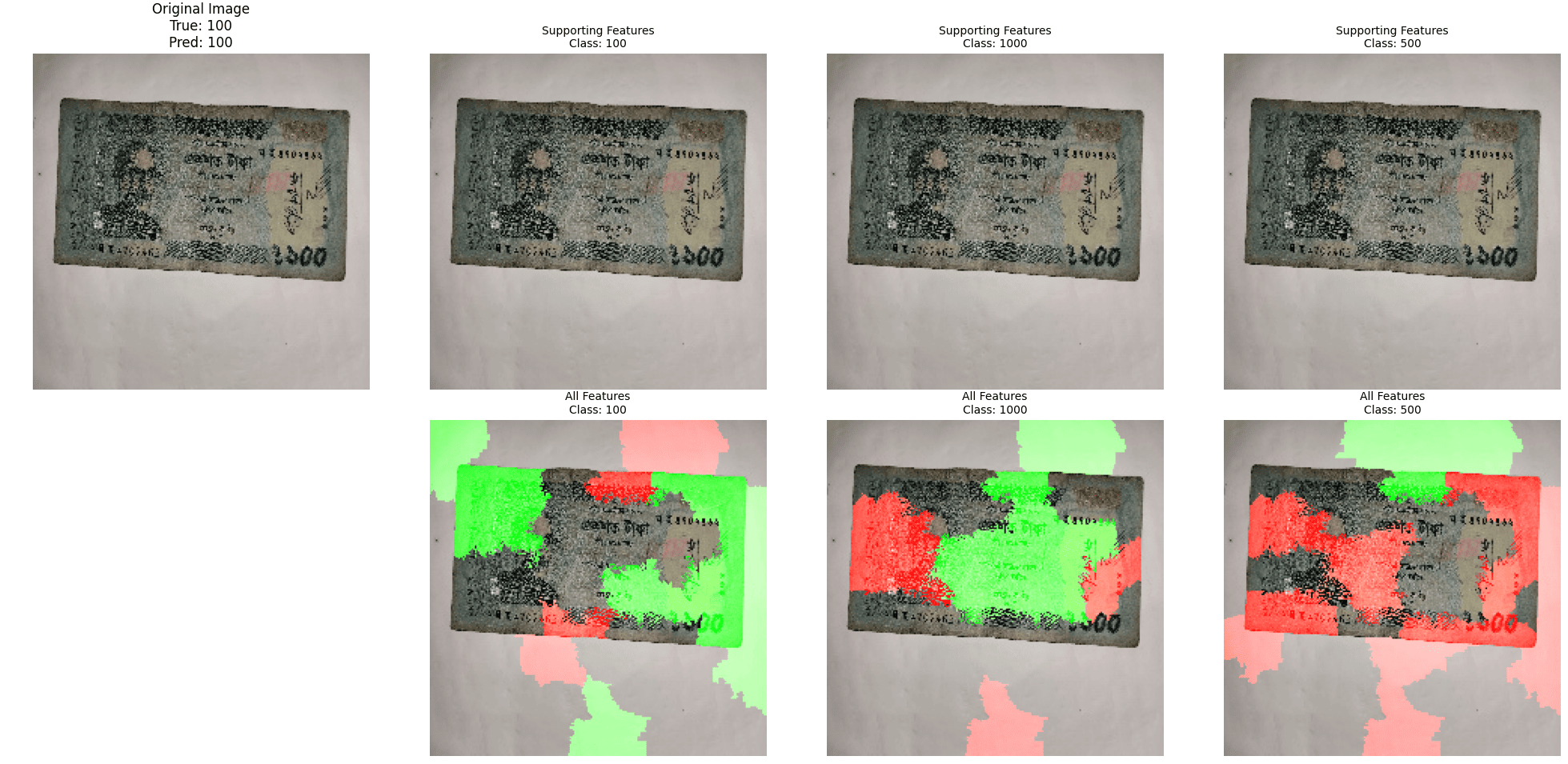}
    \hfill
    \includegraphics[width=0.32\linewidth]{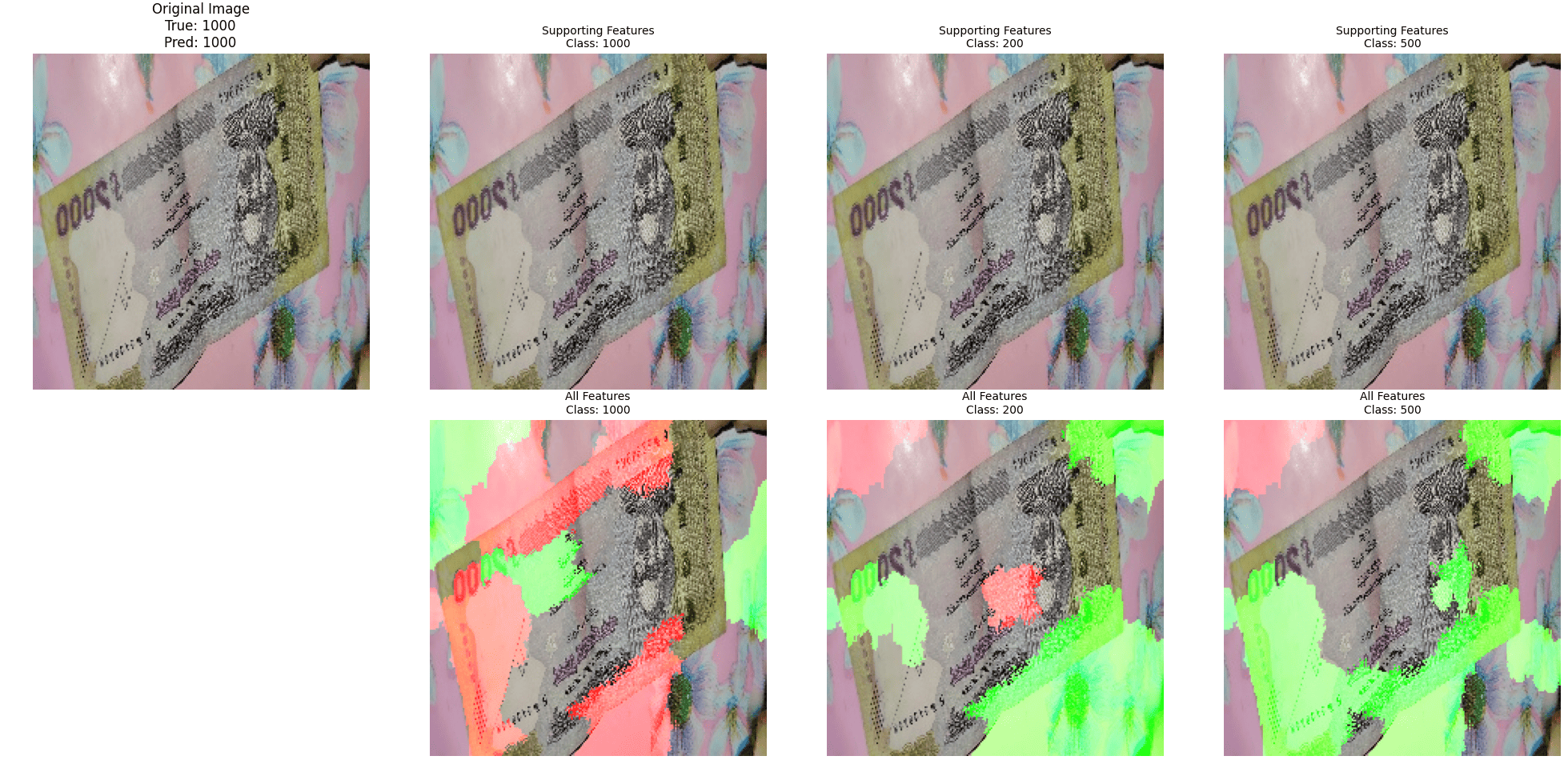}
    
    \vspace{0.4cm}
    
    \centering
    \includegraphics[width=0.32\linewidth]{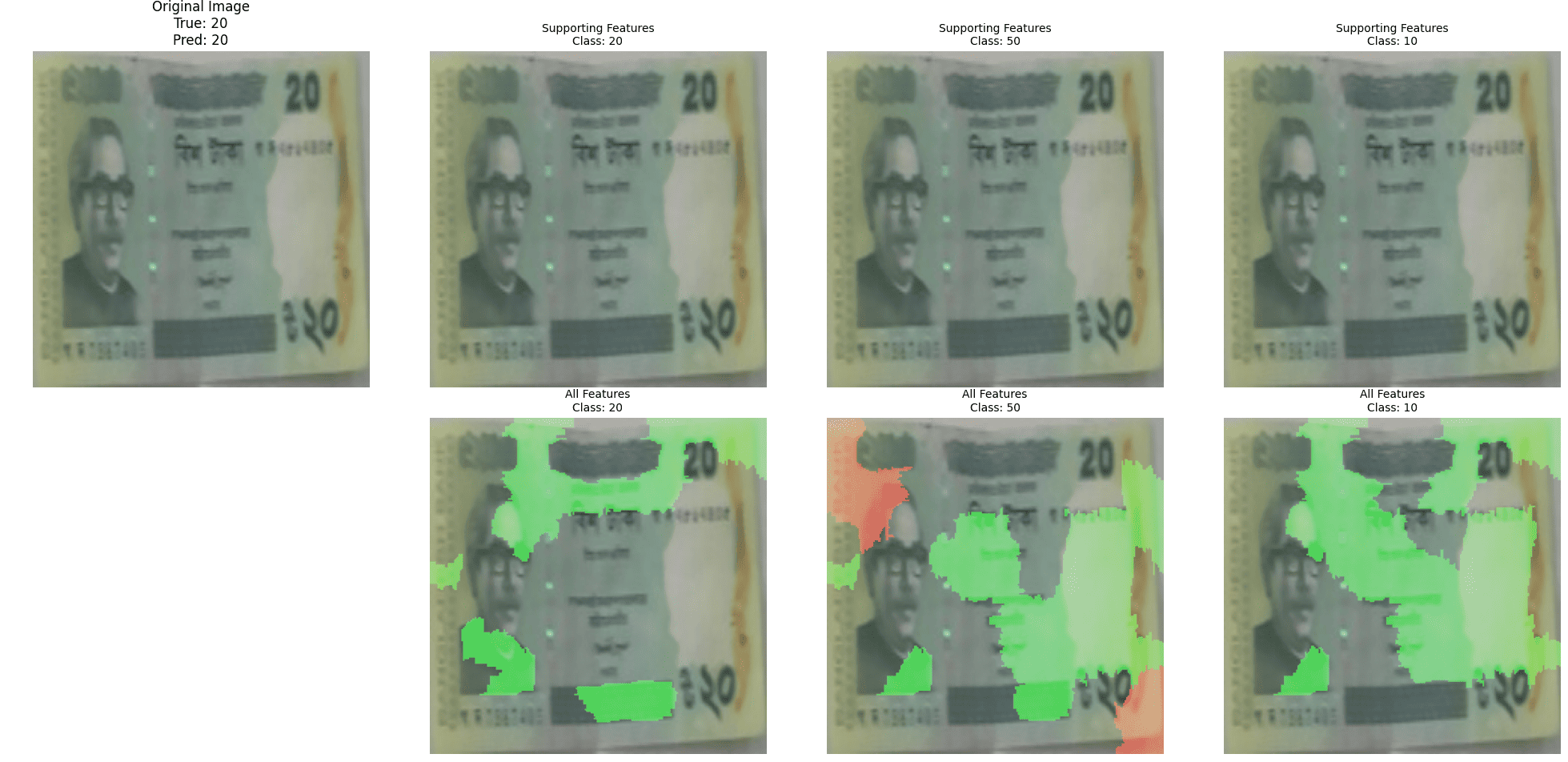}
    \hspace{0.02\linewidth}
    \includegraphics[width=0.32\linewidth]{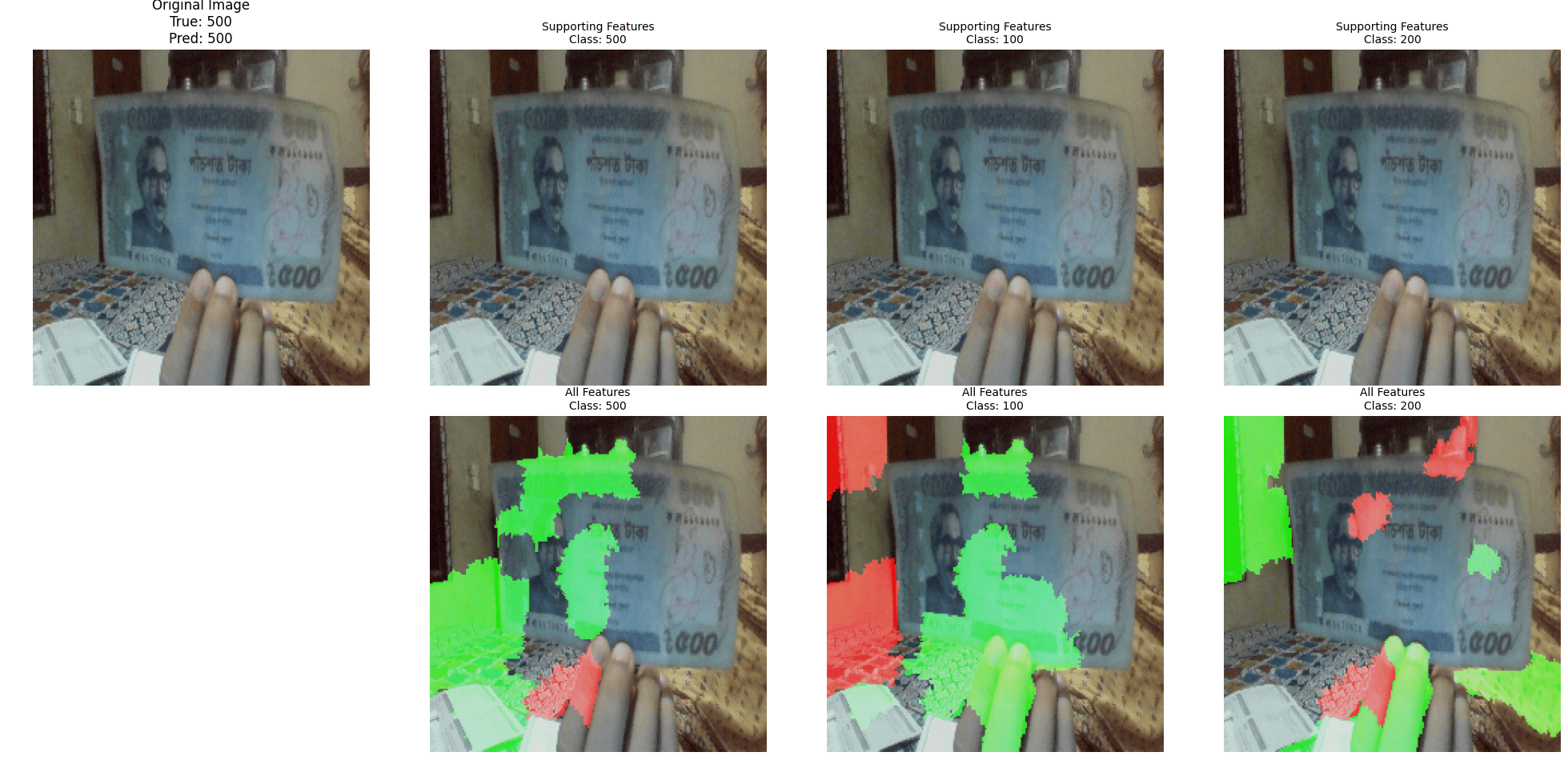}
    
    \caption{LIME-based visual explanations across all datasets. Top row: Custom Dataset-1 (200 Taka), Primary Dataset-1 (100 Taka), and Primary Dataset-2 (1000 Taka). Bottom row: Custom Dataset-2 (20 Taka) and Custom Dataset-3 (500 Taka). Green regions represent supporting features; red regions indicate contradictory evidence.}
    \label{fig:lime_all_datasets}
\end{figure*}

The LIME-based visual explanations for individual datasets are demonstrated in Figure~\ref{fig:lime_all_datasets}. The proposed model largely neglects background-interfering factors and focuses on visual information that is prominent in a denomination, including text units, colour texture patterns, numeral symbols over salted data sections, and watermarked features. More spread-out explanatory areas are observed in the higher complexity datasets, such as \textbf{Primary Dataset-2}, due to the challenge of cluttered data feature extraction. Misidentification is due to background noise and illumination, or to similarities between values that correspond to visual patterns. The model relies on large, semantically meaningful areas for classification in general, as identified by the LIME analysis.

\subsubsection{Global Feature Attribution via SHAP}

\begin{figure*}[htbp]
\centering
\includegraphics[width=0.32\linewidth,height=0.2\textheight]{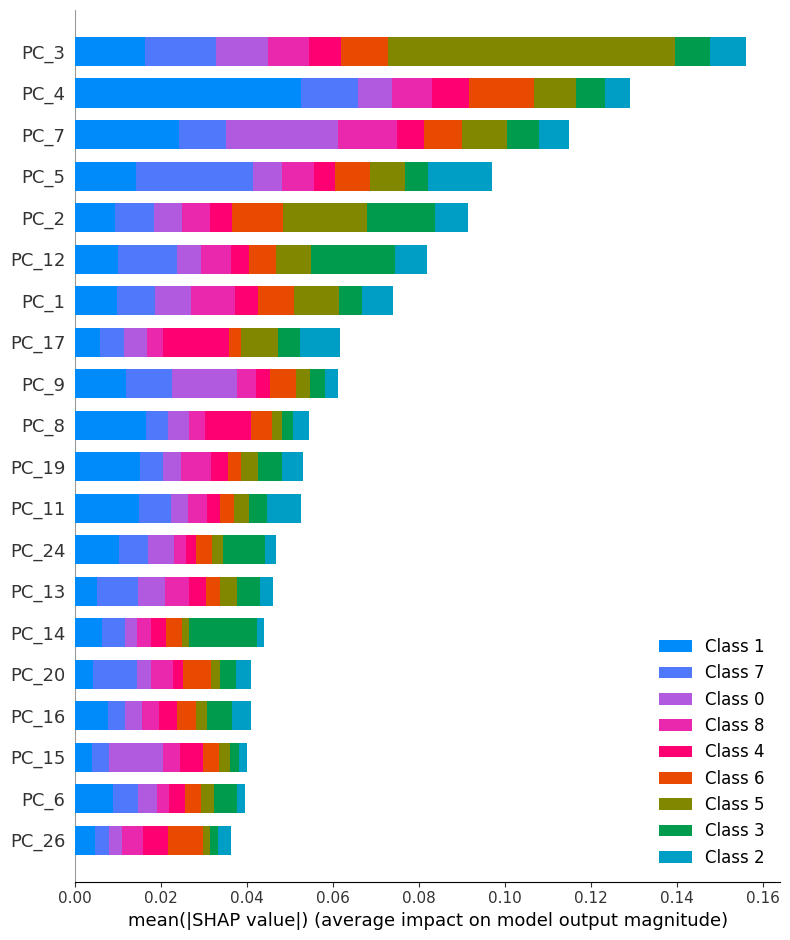}
\hfill
\includegraphics[width=0.32\linewidth,height=0.2\textheight]{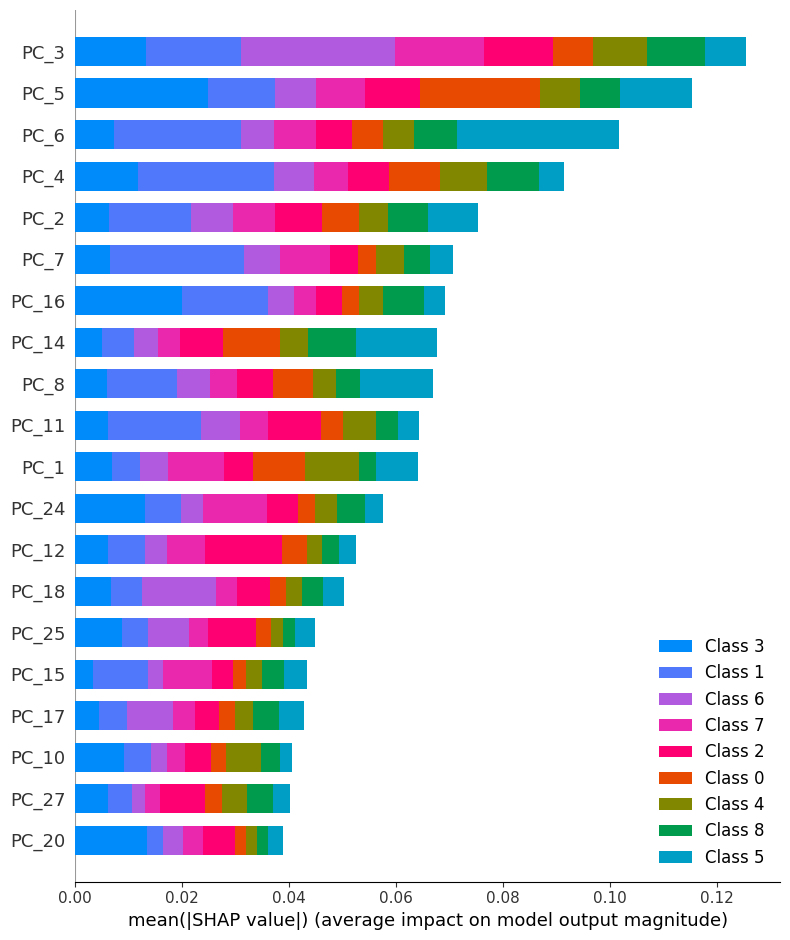}
\hfill
\includegraphics[width=0.32\linewidth,height=0.2\textheight]{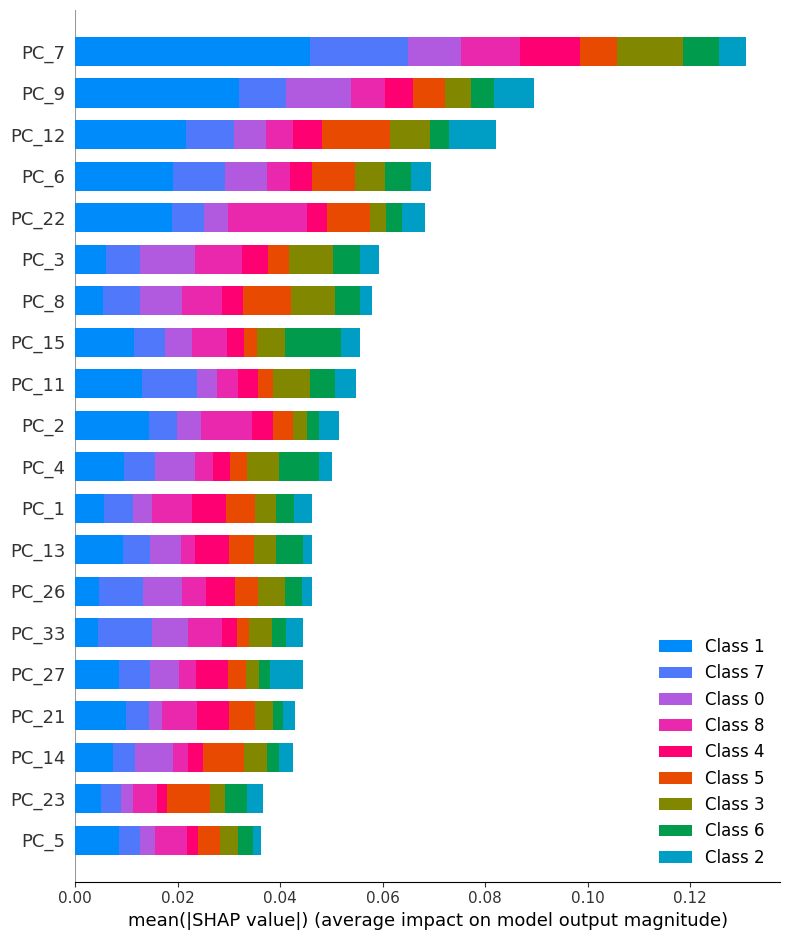}
\\[0.5cm]
\includegraphics[width=0.32\linewidth,height=0.2\textheight]{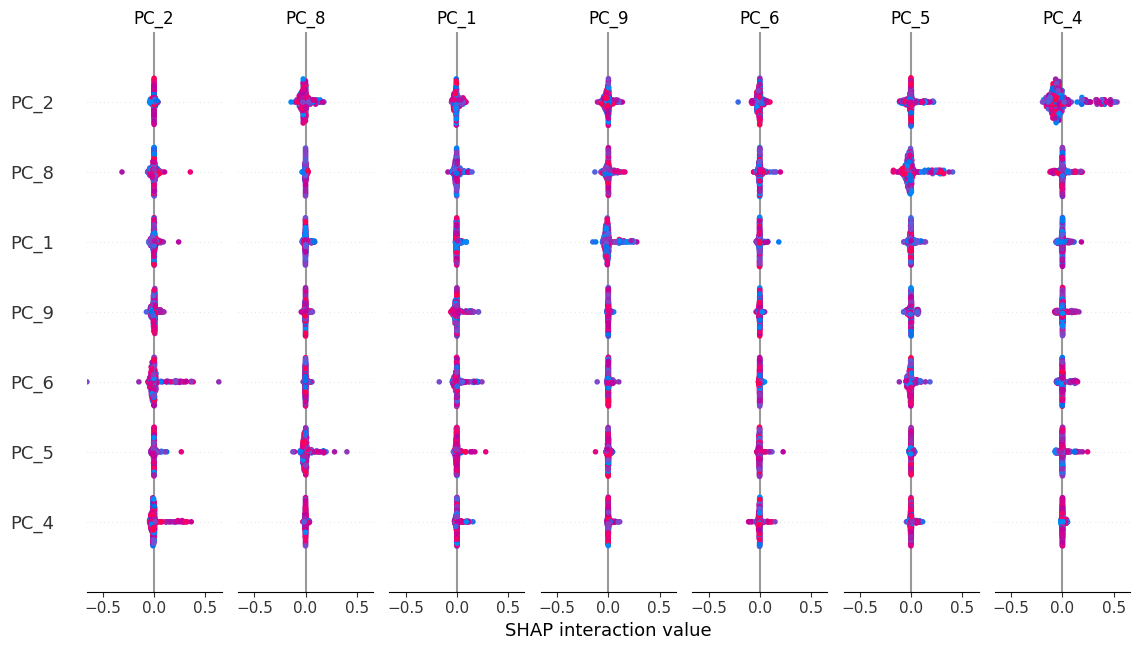}
\hfill
\includegraphics[width=0.32\linewidth,height=0.2\textheight]{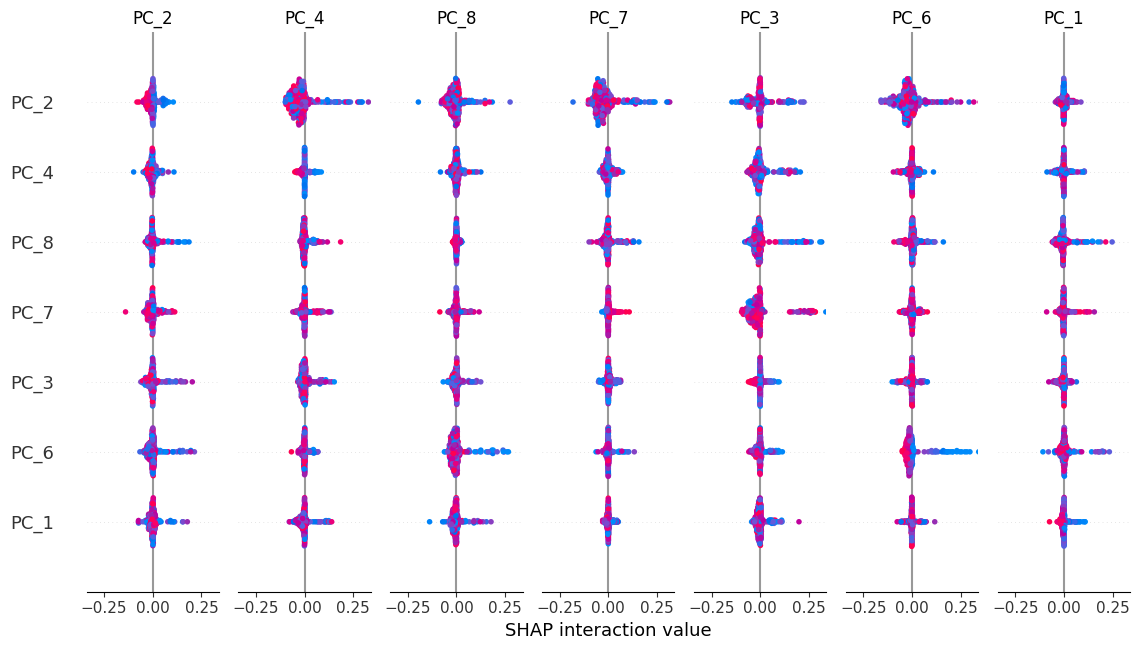}
\hfill
\includegraphics[width=0.32\linewidth,height=0.2\textheight]{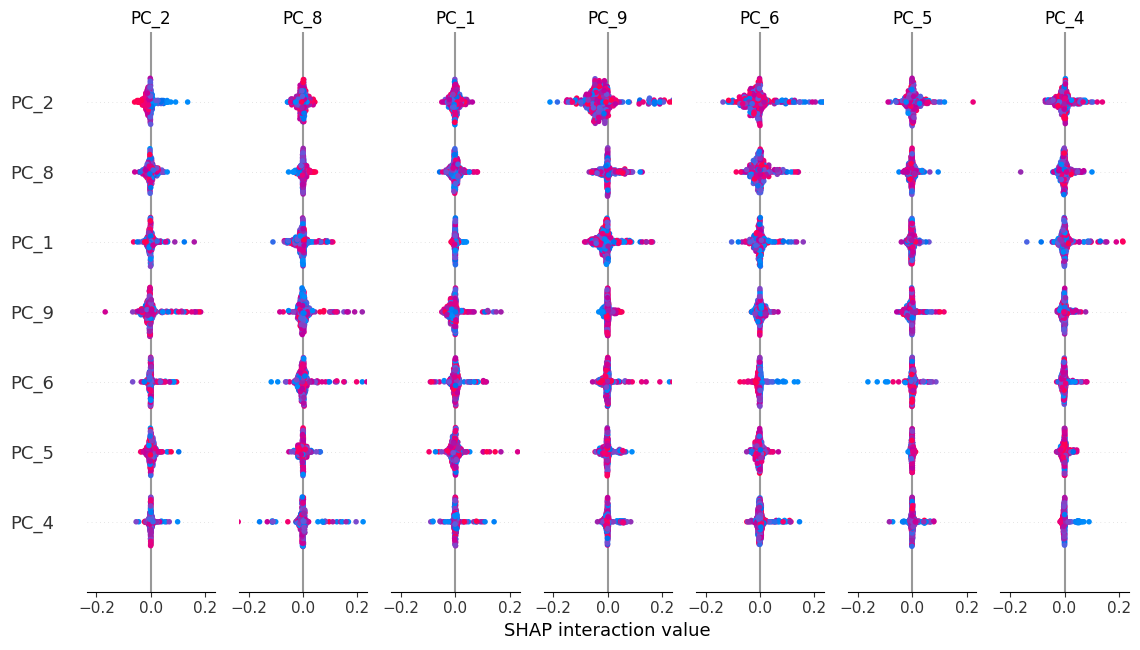}
\caption{Global SHAP analysis across Custom Dataset-1, Primary Dataset-1, and Primary Dataset-2, where SHAP summary plots (top) illustrate feature importance and impact distribution, while the SHAP dependence plots (bottom) highlight how individual features affect model predictions.}
\label{fig:shap_all_datasets}
\end{figure*}

Figure~\ref{fig:shap_all_datasets} indicates attribution of global features in SHAP over datasets. A few principal components drive most of the model’s decisions in every primary dataset. Although the dominant components differ among the datasets, their contributions correspond with the structural and color peculiarities defined by denominations. Primary components retain clear influence even in visually complex datasets, which implies that features do not rely on spurious correlations but depend on primary components.  For example, in \textbf{Primary Dataset-1}, PC\_3 contributes the most when it comes to separating the classes. A few other secondary components aid in class prediction, but only for certain types. Even in the more complicated \textbf{Primary Dataset-2}, the main components line up with clear stroke and color features. These results imply the interpretability and strength of the acquired representations under varying imaging conditions.

The findings shown in Figure~\ref{fig:lime_all_datasets} and \ref{fig:shap_all_datasets} confirm that the model learns a highly discriminative feature space and can produce interpretable and reliable predictions.


\begin{figure*}[t]
    \centering
    \includegraphics[width=\textwidth]{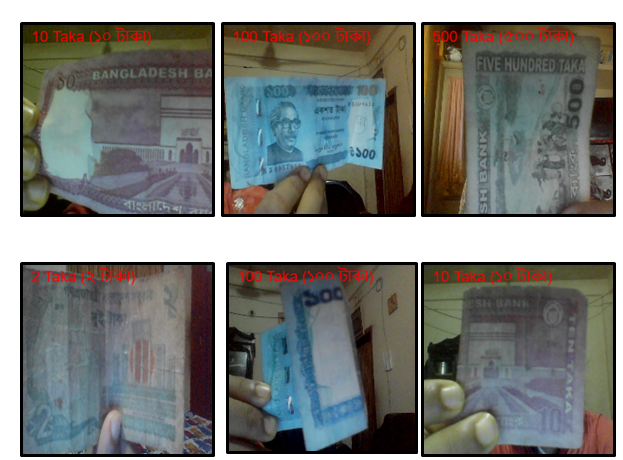}
    \caption{Real-time currency detection results showing successful identification of 2, 10, 100, and 500 Taka denominations under various lighting and orientation conditions.}
    \label{fig:realtime_results}
\end{figure*}

\subsection{Performance of the model in real-time}

The proposed model was deployed as a real-time web application on a standard laptop to demonstrate its practical usability, simulating operation on low-resource devices. The web application can be triggered by voice command and delivers prediction results in both English and Bangla, with auditory and textual outputs, enabling accessibility for visually impaired individuals. The system allows a user to click a picture of a banknote, while holding the banknote in front of the webcam and using a voice command (\"click\"), which is then automatically resized to 224x224 pixels. Figure~\ref{fig:realtime_results} demonstrates some sample predictions of Bangladeshi banknotes. These recognition results show the model's suitability for real-time deployment on resource-constrained devices.

\subsection{Comparative Performance Analysis}
\label{subsec:comparative_performance}

The performance of the proposed hybrid architecture was compared to six top models, MobileNetV2, MobileNetV3, EfficientNetB0, ResNet50V2, ResNet152V2, and InceptionV3, across the five datasets. We evaluated their performance based on Accuracy, Precision, Recall, and F1-score as shown in Table~\ref{tab:comparison_performance}. The proposed model outperformed all baseline architectures by consistently achieving better performance. Performance ranged from 92.84\% accuracy on \textbf{Primary Dataset-2} to 99.75\% on \textbf{Custom Dataset-1}, whereas precision varied from 92.88\% to 99.76\%, recall from 92.84\% to 99.75\%, and F1-scores from 92.81\% to 99.75\%.

\textbf{Primary Dataset-3} recorded the most substantial gains, with 92.84\% accuracy—a 14.15 percentage point improvement over EfficientNetB0 (78.69\%)—and 14.06 percentage point F1-score improvement (92.81\% vs 78.75\%). The model demonstrated 98.00\% precision (15.71 points over MobileNetV2) and 97.95\% recall (23.57 points over MobileNetV3) on \textbf{Primary Dataset-2}. On the \textbf{Custom Dataset-2}, the proposed model surpassed EfficientNetB0 by yielding 94.98\% accuracy, which is 8.35 percentage points higher. EfficientNetB0 showed the most consistent baseline performance and achieved second-best results on four of five datasets. Lower performance was exhibited by  ResNet architectures, where ResNet50V2 obtained only 51.76\% accuracy on \textbf{Custom Dataset-2}, which indicated generalization difficulties. These findings verify the stability of the proposed architecture's superior feature extraction followed by a classification mechanism across diverse imaging conditions to support practical financial authentication applications.

\begin{table*}[htbp]
\centering
\caption{Comparative Performance Between Baseline Models and Proposed Models Across Different Datasets (\%)}
\label{tab:comparison_performance}
\small
\setlength{\tabcolsep}{2pt}
\resizebox{\textwidth}{!}{%
\begin{tabular}{l|cccc|cccc|cccc|cccc|cccc}
\hline
\multirow{2}{*}{\textbf{Model}} 
& \multicolumn{4}{c|}{\textbf{Custom Dataset-1}} 
& \multicolumn{4}{c|}{\textbf{Primary Dataset-1}} 
& \multicolumn{4}{c|}{\textbf{Custom Dataset-2}} 
& \multicolumn{4}{c|}{\textbf{Primary Dataset-2}} 
& \multicolumn{4}{c}{\textbf{Custom Dataset-3}} \\ \cline{2-21}
 & Acc & Pre & Rec & F1 & Acc & Pre & Rec & F1 & Acc & Pre & Rec & F1 & Acc & Pre & Rec & F1 & Acc & Pre & Rec & F1 \\
\hline
MobileNetV2 & 90.45 & 76.96 & 81.51 & 79.68 & 57.64 & 82.29 & 71.80 & 72.33 & 85.81 & 88.48 & 83.55 & 82.71 & 68.85 & 82.38 & 76.20 & 76.28 & 75.88 & 82.58 & 79.13 & 79.28 \\
MobileNetV3 & 90.15 & 90.15 & 90.15 & 89.53 & 76.99 & 76.99 & 74.38 & 73.09 & 91.21 & 91.63 & 91.21 & 93.65 & 77.04 & 76.36 & 77.04 & 78.75 & 85.00 & 85.36 & 85.00 & 86.62 \\
EfficientNetB0 & 89.85 & 89.85 & 89.85 & 88.57 & 76.35 & 80.47 & 70.35 & 75.53 & 93.69 & 94.02 & 93.69 & 93.65 & 78.69 & 79.47 & 78.69 & 78.75 & 86.63 & 86.94 & 86.63 & 86.62 \\
ResNet50V2 & 72.07 & 74.33 & 72.07 & 71.91 & 49.26 & 63.73 & 49.26 & 50.95 & 65.61 & 72.89 & 65.61 & 66.26 & 45.08 & 50.54 & 46.08 & 43.68 & 51.76 & 56.73 & 51.76 & 52.03 \\
ResNet152V2 & 83.58 & 85.22 & 80.83 & 80.23 & 64.53 & 69.32 & 65.10 & 65.71 & 90.77 & 91.45 & 89.76 & 90.08 & 59.02 & 58.20 & 56.21 & 55.52 & 66.67 & 67.37 & 66.67 & 66.85 \\
InceptionV3 & 70.15 & 78.03 & 67.45 & 66.82 & 65.52 & 70.68 & 66.74 & 64.08 & 72.52 & 77.78 & 71.49 & 72.14 & 57.79 & 53.29 & 53.96 & 51.88 & 69.11 & 70.75 & 69.11 & 68.79 \\
\hline
Proposed (Hybrid) & \cellcolor{lightblue}\textcolor{skyblue}{\textbf{99.75}} & \cellcolor{lightblue}\textcolor{skyblue}{\textbf{99.76}} & \cellcolor{lightblue}\textcolor{skyblue}{\textbf{99.75}} & \cellcolor{lightblue}\textcolor{skyblue}{\textbf{99.75}} & \cellcolor{lightblue}\textcolor{skyblue}{\textbf{97.95}} & \cellcolor{lightblue}\textcolor{skyblue}{\textbf{98.00}} & \cellcolor{lightblue}\textcolor{skyblue}{\textbf{97.95}} & \cellcolor{lightblue}\textcolor{skyblue}{\textbf{97.96}} & \cellcolor{lightblue}\textcolor{skyblue}{\textbf{98.66}} & \cellcolor{lightblue}\textcolor{skyblue}{\textbf{98.69}} & \cellcolor{lightblue}\textcolor{skyblue}{\textbf{98.66}} & \cellcolor{lightblue}\textcolor{skyblue}{\textbf{98.66}} & \cellcolor{lightblue}\textcolor{skyblue}{\textbf{92.84}} & \cellcolor{lightblue}\textcolor{skyblue}{\textbf{92.88}} & \cellcolor{lightblue}\textcolor{skyblue}{\textbf{92.84}} & \cellcolor{lightblue}\textcolor{skyblue}{\textbf{92.81}} & \cellcolor{lightblue}\textcolor{skyblue}{\textbf{94.98}} & \cellcolor{lightblue}\textcolor{skyblue}{\textbf{94.97}} & \cellcolor{lightblue}\textcolor{skyblue}{\textbf{94.98}} & \cellcolor{lightblue}\textcolor{skyblue}{\textbf{94.96}} \\
\hline
\end{tabular}%
}
\vspace{0.1cm}
\small
\\ \textbf{Abbreviations:} Acc=Accuracy, Prec=Precision, Rec=Recall, F1=F1-score
\vspace{0.1cm}
\end{table*}

\section{Conclusion}
\label{section:conclusion}
This study presents a hybrid CNN framework for Bangladeshi banknote recognition to aid the visually impaired individuals, combining multiple state-of-the-art architectures to achieve an accurate, real-time system. Extensive evaluations demonstrate its robustness across diverse imaging conditions, outperforming individual models in handling complex backgrounds. Future endeavors include expanding the dataset to incorporate newly issued banknotes and coins, integrating counterfeit detection mechanisms such as OCR, and optimizing the model for deployment on mobile and edge devices. The proposed approach offers a practical solution to assist visually impaired individuals in financial transactions, providing both accuracy and efficiency while ensuring financial independence.

\bibliographystyle{IEEEtran}

\begin{thebibliography}{100}
\providecommand{\url}[1]{#1}
\csname url@samestyle\endcsname
\providecommand{\newblock}{\relax}
\providecommand{\bibinfo}[2]{#2}
\providecommand{\BIBentrySTDinterwordspacing}{\spaceskip=0pt\relax}
\providecommand{\BIBentryALTinterwordstretchfactor}{4}
\providecommand{\BIBentryALTinterwordspacing}{\spaceskip=\fontdimen2\font plus
\BIBentryALTinterwordstretchfactor\fontdimen3\font minus \fontdimen4\font\relax}
\providecommand{\BIBforeignlanguage}[2]{{%
\expandafter\ifx\csname l@#1\endcsname\relax
\typeout{** WARNING: IEEEtran.bst: No hyphenation pattern has been}%
\typeout{** loaded for the language `#1'. Using the pattern for}%
\typeout{** the default language instead.}%
\else
\language=\csname l@#1\endcsname
\fi
#2}}
\providecommand{\BIBdecl}{\relax}
\BIBdecl



\bibitem{1kola2023life}
Š. Kolačko, J. Predović, A. Tomić, and V. Oršulić, ``Life quality in patients with impaired visual acuity undergoing intravitreal medication applications,'' \textit{Int. J. Environ. Res. Public Health}, vol. 20, no. 4, 2023. [Online]. Available: https://www.mdpi.com/1660-4601/20/4/2879

\bibitem{VIPassist1}
M. D. Messaoudi, B.-A. J. Menelas, and H. Mcheick, ``Review of navigation assistive tools and technologies for the visually impaired,'' \textit{Sensors}, vol. 22, no. 20, 2022. [Online]. Available: https://www.mdpi.com/1424-8220/22/20/7888

\bibitem{VIPassist2}
S. Zafar, M. Asif, M. B. Ahmad, T. M. Ghazal, T. Faiz, M. Ahmad, and M. A. Khan, ``Assistive devices analysis for visually impaired persons: A review on taxonomy,'' \textit{IEEE Access}, vol. 10, pp. 13354--13366, 2022.

\bibitem{3world_2023}
World Health Organization, ``Blindness and vision impairment,'' Aug. 2023. [Online]. Available: https://www.who.int/news-room/fact-sheets/detail/blindness-and-visual-impairment

\bibitem{bd_blindness2003prevalence}
B. Dineen, R. Bourne, S. Ali, D. N. Huq, and G. Johnson, ``Prevalence and causes of blindness and visual impairment in Bangladeshi adults: Results of the national blindness and low vision survey of Bangladesh,'' \textit{Brit. J. Ophthalmol.}, vol. 87, no. 7, pp. 820--828, 2003.

\bibitem{bd_blindness2022prevalence}
S. A. Shakoor, M. Rahman, A. E. Hossain, M. Moniruzzaman, M. R. Bhuiyan, F. Hakim, and M. M. Zaman, ``Prevalence of blindness and its determinants in Bangladeshi adult population: Results from a national cross-sectional survey,'' \textit{BMJ Open}, vol. 12, no. 4, p. e052247, 2022.

\bibitem{intaglio_tactile}
M. S. Uddin, P. P. Das, and M. S. A. Roney, ``Image-based approach for the detection of counterfeit banknotes of Bangladesh,'' in \textit{Proc. 5th Int. Conf. Informatics, Electron. Vis. (ICIEV)}, 2016, pp. 1067--1072.

\bibitem{bd_tactile_wiki}
C. to, ``Bangladeshi 10-taka banknote,'' Feb. 2025. [Online]. Available: https://en.wikipedia.org/wiki/Bangladeshi\_10-taka\_note

\bibitem{bd_tactile_web}
K. Platform and K. Platform, ``Bangladesh releases the first three banknotes in new series - Keesing Platform,'' Jun. 2025. [Online]. Available: https://platform.keesingtechnologies.com/bangladesh-releases-the-first-three-banknotes-in-new-series/

\bibitem{tactile_semary2015currency}
N.~Semary, S.~Fadl, M.~Eissa, and A.~Gad, ``Currency recognition system for visually impaired: Egyptian banknote as a study case,'' in \emph{Proc. IEEE Int. Conf. Inf. Commun. Technol. Accessibility (ICTA)}, Dec. 2015, doi: 10.1109/ICTA.2015.7426896.


\bibitem{tactile_yousry2018currency}
A. Yousry, M. Taha, and M. M. Selim, ``Currency recognition system for blind people using ORB algorithm,'' \textit{Int. Arab. J. e Technol.}, vol. 5, no. 1, pp. 34--40, 2018.

\bibitem{MobileNetV3}
S. Qian, C. Ning, and Y. Hu, ``MobileNetV3 for image classification,'' in \textit{Proc. IEEE 2nd Int. Conf. Big Data, Artif. Intell. Internet Things Eng. (ICBAIE)}, 2021, pp. 490--497. [Online]. Available: https://api.semanticscholar.org/CorpusID:233177580

\bibitem{tan2019efficientnet}
M. Tan and Q. V. Le, ``EfficientNet: Rethinking model scaling for convolutional neural networks,'' in \textit{Proc. Int. Conf. Mach. Learn. (ICML)}. PMLR, 2019, pp. 6105--6114.

\bibitem{9ganjavecurrency}
P. Ganjave, R. Markad, G. Rasal, and Y. Kalekar, ``Currency detector for visually impaired (study of the system which identifies Indian currency for blind people),'' Dept. Comput. Eng., Pimpri Chinchwad College Eng., Pune, India, 2020.


\bibitem{10roy2017development}
S. Roy, S. Banerjee, K. Chowdhury, and U. Biswas, ``Development and analysis of a three phase cloudlet allocation algorithm,'' \textit{J. King Saud Univ.-Comput. Inf. Sci.}, vol. 29, no. 4, pp. 473--483, 2017.

\bibitem{Sarker2019}
M. F. R. Sarker, M. I. M. Raju, A. A. Marouf, R. Hafiz, S. A. Hossain, and M. H. K. Protik, ``Real-time Bangladeshi currency detection system for visually impaired person,'' in \textit{Proc. Int. Conf. Bangla Speech Lang. Process. (ICBSLP)}, 2019, pp. 1--4.


\bibitem{12linkon2020deep}
A. H. M. Linkon, M. M. Labib, F. H. Bappy, S. Sarker, M.-E. Jannat, and M. S. Islam, ``Deep learning approach combining lightweight CNN architecture with transfer learning: An automatic approach for the detection and recognition of Bangladeshi banknotes,'' in \textit{Proc. 11th Int. Conf. Elect. Comput. Eng. (ICECE)}. IEEE, 2020, pp. 214--217.

\bibitem{13joshi2020yolov3}
R. C. Joshi, S. Yadav, and M. K. Dutta, ``YOLO-v3 based currency detection and recognition system for visually impaired persons,'' in \textit{Proc. Int. Conf. Contemp. Comput. Appl. (IC3A)}, 2020, pp. 280--285.

\bibitem{8singh2022ipcrf}
M. Singh, J. Chauhan, M. S. Kanroo, S. Verma, and P. Goyal, ``IPCRF: An end-to-end Indian paper currency recognition framework for blind and visually impaired people,'' \textit{IEEE Access}, vol. 10, pp. 90726--90744, 2022.

\bibitem{MobileApp1Pujari}
V. Pujari, K. Madnal, and D. Premchandran, ``Mobile app for enhancing accessibility among the visually impaired,'' in \textit{Proc. 2nd DMIHER Int. Conf. Artif. Intell. Healthcare, Educ. Ind. (IDICAIEI)}, 2024, pp. 1--6.

\bibitem{hybrid1_nazir2024deep}
A. Nazir, J. He, N. Zhu, S. S. Qureshi, S. U. Qureshi, F. Ullah, A. Wajahat, and M. S. Pathan, ``A deep learning-based novel hybrid CNN-LSTM architecture for efficient detection of threats in the IoT ecosystem,'' \textit{Ain Shams Eng. J.}, vol. 15, no. 7, p. 102777, 2024.

\bibitem{hybrid2_yang2025enhanced}
J. Yang, H. Wan, and Z. Shang, ``Enhanced hybrid CNN and transformer network for remote sensing image change detection,'' \textit{Sci. Rep.}, vol. 15, no. 1, p. 10161, 2025.

\bibitem{hybrid3_terzioglu2024optimizing}
F. Terzioglu, E. Y. Sidky, J. P. Phillips, I. S. Reiser, G. Bal, and X. Pan, ``Optimizing dual-energy CT technique for iodine-based contrast-to-noise ratio, a theoretical study,'' \textit{Med. Phys.}, vol. 51, no. 4, pp. 2871--2881, 2024.

\bibitem{hybrid4_ye2025hybrid}
Y. Ye, K. Chipusu, M. A. Ashraf, B. Ding, Y. Huang, and J. Huang, ``Hybrid CNN-BLSTM architecture for classification and detection of arrhythmia in ECG signals,'' \textit{Sci. Rep.}, vol. 15, no. 1, p. 34510, 2025.

\bibitem{7tasnim2021bangladeshi}
R. Tasnim, S. T. Pritha, A. Das, and A. Dey, ``Bangladeshi banknote recognition in real-time using convolutional neural network for visually impaired people,'' in \textit{Proc. 2nd Int. Conf. Robot., Elect. Signal Process. Techn. (ICREST)}. IEEE, 2021, pp. 388--393.

\bibitem{xai_ribeiro2016should}
M. T. Ribeiro, S. Singh, and C. Guestrin, ``Why should I trust you? Explaining the predictions of any classifier,'' in \textit{Proc. 22nd ACM SIGKDD Int. Conf. Knowl. Discovery Data Mining}, 2016, pp. 1135--1144.

\bibitem{xai_lundberg2017unified}
S. M. Lundberg and S.-I. Lee, ``A unified approach to interpreting model predictions,'' \textit{Adv. Neural Inf. Process. Syst.}, vol. 30, 2017.

\bibitem{Akter2018}
J. Akter, M. K. Hossen, and M. S. A. Chowdhury, ``Bangladeshi paper currency recognition system using supervised learning,'' in \textit{Proc. Int. Conf. Comput., Commun., Chem., Mater. Electron. Eng. (IC4ME2)}, 2018, pp. 1--4.



\bibitem{Banglacurrency2019}
H. Murad, N. I. Tripto, and M. E. Ali, ``Developing a bangla currency recognizer for visually impaired people,'' in \textit{Proc. 10th Int. Conf. Inf. Commun. Technol. Develop. (ICTD)}, ser. ICTD '19. New York, NY, USA: Association for Computing Machinery, 2019. [Online]. Available: https://doi.org/10.1145/3287098.3287152

\bibitem{BanglaMoney2018}
N. Sojib, ``Bangla money,'' 2018. [Online]. Available: https://www.kaggle.com/datasets/nsojib/bangla-money

\bibitem{PasumarthyYOLOV52022}
M. Pasumarthy, R. Padhy, R. Yadav, G. Subramaniam, and M. Rao, ``An Indian currency recognition model for assisting visually impaired individuals,'' in \textit{Proc. IEEE Int. Conf. Recent Adv. Syst. Sci. Eng. (RASSE)}, 2022, pp. 1--5.

\bibitem{200tk_release}
C. to, ``Banknotes of the Bangladeshi taka,'' Jun. 2025. [Online]. Available: https://en.wikipedia.org/wiki/Banknotes\_of\_the\_Bangladeshi\_taka

\bibitem{BanglaTaka2025}
M. N. I. Nuhash and S. Akter, ``BanglaTaka: A dataset for classification of Bangladeshi banknotes,'' \textit{Data Brief}, vol. 61, p. 111853, 2025. [Online]. Available: https://www.sciencedirect.com/science/article/pii/S2352340925005785

\bibitem{Siddiki2023}
M. A. Siddiki, A. B. Siddique, J. Hossain, M. M. Rahman, and M. S. Rahman, ``Bangladeshi currency identification and fraudulence detection using deep learning and feature extraction,'' \textit{Int. J. Comput. Sci. Mobile Comput.}, vol. 12, no. 1, pp. 1--13, 2023.

\bibitem{Chowdhury2024}
M. E. A. Chowdhury, S. I. Islam, T. Alam, and S. Ahmed, ``Classification of Bangladeshi currency using convolutional neural network in cross-dataset recognition environment,'' in \textit{Proc. 6th Int. Conf. Elect. Eng. Inf. Commun. Technol. (ICEEICT)}, 2024, pp. 296--301.


\bibitem{Das2023CNN}
D. Das, D. Kundu, S. Sazzad, and A. Rahman, ``Utilizing deep learning with CNN model for precise identification of Bangladeshi currency notes to aid visually impaired people,'' in \textit{Proc. 5th Int. Conf. Sustain. Technol. Ind. 5.0 (STI)}, 2023, pp. 1--6.

\bibitem{park2023mbdm}
C. Park and K. R. Park, ``MBDM: Multinational banknote detecting model for assisting visually impaired people,'' \textit{Mathematics}, vol. 11, no. 6, p. 1392, 2023.

\bibitem{Nasir2024}
A. Nasir, S. Zafar, and Z. Nasir, ``Pakistani currency recognition and classification for visually impaired people using convolutional neural network,'' in \textit{Proc. 4th Int. Conf. Digit. Futures Transform. Technol. (ICoDT2)}, 2024, pp. 1--6.

\bibitem{14IslamViT}
M. F. Islam, J. Hasan, and M. S. Rahman, ``Bangladeshi paper currency recognition with improved dataset using vision transformer,'' in \textit{Proc. 6th Int. Conf. Elect. Eng. Inf. Commun. Technol. (ICEEICT)}, 2024, pp. 226--229.

\bibitem{AugmentedBanglaMoney}
T. Mohammed, ``Augmented Bangla money dataset with 200 taka,'' 2022. [Online]. Available: https://www.kaggle.com/datasets/tazwarmohammed/augmented-bangla-money-dataset-with-200-taka

\bibitem{Ali2024Egyptian}
H. R. Ali, Y. M. Elgamel, and M. A. Mohamed, ``Recognizing Egyptian currency for people with visual impairment using deep learning models,'' \textit{Sci. Rep.}, vol. 14, no. 1, p. 20646, 2024.

\bibitem{Evwiekpaefe2024}
A. E. Evwiekpaefe and H. Isacha, ``A visually impaired mobile application for currency recognition using MobileNetV2 CNN architecture,'' \textit{Int. J. Adv. Res. Publ. Rev.}, vol. 2, no. 9, pp. 725--732, 2024.

\bibitem{Neto2023Brazilian}
O. L. S. Neto, F. G. Oliveira, J. M. B. Cavalcanti, and J. L. S. Pio, ``Brazilian banknote recognition based on CNN for blind people,'' in \textit{Proc. 18th Int. Joint Conf. Comput. Vis., Imag. Comput. Graph. Theory Appl. (VISIGRAPP)}, 2023, pp. 804--811.

\bibitem{Awad2022Iraqi}
S. R. Awad, B. T. Sharef, A. M. Salih, and F. L. Malallah, ``Deep learning-based Iraqi banknotes classification system for blind people,'' \textit{Eastern-Eur. J. Enterprise Technol.}, vol. 1, no. 2, pp. 31--38, 2022.

\bibitem{Bolanos2024Colombian}
C. Bolaños-Fernández and E. B. Bacca-Cortes, ``Mobile application for recognizing Colombian currency with audio feedback for visually impaired people,'' \textit{Ingeniería}, vol. 29, no. 2, p. e20743, 2024.

\bibitem{Howard2019mobv3}
A. Howard, M. Sandler, G. Chu, L.-C. Chen, B. Chen, M. Tan, W. Wang, Y. Zhu, R. Pang, V. Vasudevan \textit{et al.}, ``Searching for MobileNetV3,'' in \textit{Proc. IEEE/CVF Int. Conf. Comput. Vis. (ICCV)}, 2019, pp. 1314--1324.
\end{thebibliography}

\end{document}